\documentclass[preprint,review,12pt]{elsarticle}
 
  \usepackage{algorithm2e}
  \usepackage{graphics}
  \usepackage{subfig}
  \usepackage{algorithm2e}
  \usepackage{setspace}
  \usepackage{verbatim} 
  \usepackage{footnote}
  \usepackage{amssymb}
  \usepackage{amsthm}
  \usepackage{amsmath}
  \usepackage{amsfonts}
  \usepackage{amsxtra}
  \usepackage{color}
  \usepackage{amssymb, pxfonts}
  \usepackage{mathtools}
  \usepackage{setspace}
  \usepackage{url}
  \usepackage[section]{placeins} 
  \usepackage{array}    
  \usepackage{booktabs} 
  \usepackage{multirow} 
  \usepackage{lineno}
    \usepackage{bm}
  \usepackage{sansmath}
  \usepackage{diagbox}
\usepackage{hyperref}
\usepackage{fancyhdr}

\begin{document}
    \begin{frontmatter}

  \title{\normalsize This manuscript is a preprint version. The final version of this paper is available in Engineering Applications of Artificial Intelligence, vol${.}$ 72, pp${.}$ 368-381, 2018. DOI: 10.1016/j${.}$engappai${.}$2018.04.013 \\ \vspace{1cm} \rule[1cm]{5.43in}{0.1pt}
  \large How far did we get in face spoofing detection?}

  \cortext[cor1]{Corresponding author: Luciano Oliveira, tel. +55 71 3283-9472}
  \author{Luiz Souza}\ead{luiz.otavio@ufba.br}
   \author{Luciano Oliveira}\ead{lrebouca@ufba.br}
   \author{Mauricio Pamplona}\ead{mauricio@dcc.ufba.br}
  \address{IVISION Lab, Federal University of Bahia}
\author{Joao Papa}\ead{papa@fc.unesp.br}
  \address{RECOGNA Lab, S\~{a}o Paulo State University}
  \begin{abstract}

The growing use of control access systems based on face recognition shed light over the need for even more accurate systems to detect face spoofing attacks. In this paper, an extensive analysis on face spoofing detection works published in the last decade is presented. The analyzed works are categorized by their fundamental parts, \emph{i.e.,} descriptors and classifiers. This structured survey also brings a comparative performance analysis of the works considering the most important public data sets in the field. The methodology followed in this work is particularly relevant to observe temporal evolution of the field, trends in the existing approaches, to discuss still opened issues, and to propose new perspectives for the future of face spoofing detection.
  \end{abstract}


\makeatletter
\def\ps@pprintTitle{%
   \let\@oddhead\@empty
   \let\@evenhead\@empty
   \let\@oddfoot\@empty
   \let\@evenfoot\@oddfoot
}
\makeatother

 \end{frontmatter}
 %\linenumbers
  
  \section{Introduction} \label{sec:introduction}

  In the last decade, there has been an increasing interest in human automatic secure identification, being mainly based on unique personal biometric information~\cite{jain:2008}. One of the main reasons for such focus concerns the high number of security breaches and transaction frauds in non-biometric systems, which are prone to be cracked due to inherent vulnerabilities~\cite{meadowcroft:2008}, like stolen cards and shared passwords, just to name a few.

Biometrics may use physical or behavioral characteristics for identification purposes, and different alternatives have been explored over the years: fingerprint ~\cite{hasan:2013,marasco2015asurvey,peralta2014minutiae}, hand geometry ~\cite{aleidan:2013,michael2012acontactless}, palmprint ~\cite{tamrakar2016kernel}, voice ~\cite{yadav:2013,choi2015unsupervised}, face ~\cite{zhao:2003,feng2016integration,dora2017anevolutionary}, and handwritten signature ~\cite{sanmorino:2012}. Among those, face stands out for its acceptability and recognition cost, turning out to be one of the best option for a wide range of applications, from low-security uses ({\it e.g.,} social media and smartphone access control) to high-security applications ({\it e.g.,} border control and video surveillance in critical places). 

\begin{figure}[!b]
  \centering
  \includegraphics[width=0.18\textwidth]{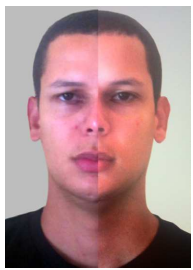} 
  \caption{Example of a half real (photo) and half fake face (photo of a photo). Which half is the real one? The answer is the one on the left.}
  \label{fig01}
  \end{figure}

This popularity, however, comes with a price: face recognition systems have become a major target of spoofing attacks. In such scenarios, an impostor attempts to be granted in an identification process by forging someone else's identity. As procedures to replicate human faces are very much standard nowadays ({\it e.g.,} photo and 3D printing), spoofing detection has become mandatory in any suitable face recognition system. Figure~\ref{fig01} illustrates the complexity of this problem, and the following question can be raised: "Which half is real or fake?". It is sometimes a very challenging task, even for humans.

Several approaches for spoofing detection have been developed in the last decade. Recently, two main surveys on the subject present a comprehensive review ~\cite{galbally:2014,parveen2015face}: in ~\cite{galbally:2014}, a survey on anti-spoofing methods focuses not only on face, but also on other biometric traits (\textit{e.g.}, iris, voice, fingerprint); in ~\cite{parveen2015face}, face anti-spoofing methods are discussed by considering the intrusiveness of each method, with few attention on comparative analysis and temporal evolution of the field. On the other hand, the proposed survey focuses only on face-oriented works, reviewing and analyzing the most relevant works on face spoofing detection in the literature towards depicting the advance of the detection methods in the last decade. An extensive set of face anti-spoofing methods is presented, also depicting the evolution of the existing works. In this sense, trends denoted throughout these years were pointed out, as well as open issues were remarked in order to provide new directions on research topics in the future. Next, the contributions of this survey are addressed and discussed in details with respect to the other existing surveys, with special attention to the gaps filled by the present work.

\subsection{Contributions}
  
To the best of our knowledge, there are only two surveys in the context of face spoofing detection~\cite{galbally:2014,parveen2015face}. Although two face anti-spoofing competitions were organized \cite{chakka:2011,chingovska:2013}, and several data sets and methods have been published, the amount of gathered data and results were not still thorough and critically analyzed so far. Even these two existing surveys do not concentrate efforts to understanding the trends of this research field in terms of conception of the methods and results. 

Galbally \textit{et al.}~\cite{galbally:2014} published a survey based on a chronological evolution of multimodal anti-spoofing methods. Although a special attention was given to face anti-spoofing, other biometric traits were also presented and discussed. A proposed timeline takes into consideration fingerprint, iris, and face anti-spoofing detection competitions, being the latter one organized by one of the authors of the survey~\cite{galbally:2014}. In regard to face-driven works, the authors provided an extensive and comprehensive description of different types of face attacks and public image data sets. The face anti-spoofing methods, categorized by Galbally \textit{et al.}, were according to three levels: sensor, features, and multi-modal fusion, but being only two levels employed to classify the analyzed works. Sixteen existing works compose the face study part, which was characterized by the level of the technique, type of attack, public image data set used, and a single error rate. At the end, a discussion was addressed showing that although competitive laboratory performances were achieved, some people were successfully able to hack the fingerprint recognition system of the Iphone 5s. In~\cite{galbally:2014}, also, some discussion about performance of face anti-spoofing methods resided in general considerations about cross-data set performance evaluation (in order to turn methods' evaluation more thoroughly accomplished), new relevant features acquired on facial blood flow, and new hardware that could be used along with cameras to improve face anti-spoofing detection. The remainder of the survey in~\cite{galbally:2014} discusses philosophical aspects of performing an anti-spoofing detection approach within face recognition systems. 

Parveen \textit{et al.}~\cite{parveen2015face} followed a general architecture comprised of a sensor, pre-processing, feature extraction, and classification steps as a basis for a taxonomy of face anti-spoofing detection methods. The methods are categorized as non-intrusive or intrusive ones, addressed according to the stillness or motion detection presented in the detection process, respectively. Twenty-nine face anti-spoofing methods were studied, and the results of the existing works were individually analyzed over public image data sets. An experimental analysis was carried out by means of four error measures: \textit{half total error rate} (HTER), \textit{equal error rate} (EER), \textit{area under curve} (AUC) and \textit{accuracy} (ACC). At the end in~\cite{parveen2015face}, some pros and cons are highlighted with regard to implementation complexity, user collaboration and attack coverage. 

Differently from Galbally \textit{et al.}~\cite{galbally:2014}, which spread out the discussion on various anti-spoofing methods using different traits, we present an extensive survey that is focused on the evolution of particularly face spoofing detection methods and existing benchmarks. Instead of following a more generic categorization as those proposed in~\cite{galbally:2014,parveen2015face}, all gathered works here were organized in terms of their main component parts, \emph{i.e.,} descriptors and classifiers (see Section \ref{sec:detection}). This taxonomy was devised to help the reader to better understand the processes behind each countermeasure, and to unveil technical trends concerning different types of attacks. Since all works comprise features and learning methods, this organization seems to be the best to depict a big picture of the state-of-the-art research related to face spoofing detection.

Despite the other two surveys, our work resorts to a quantitative and analytical methodology (see Section \ref{sec:comparative}) in order to support the analysis of trends of the existing face anti-spoofing approaches (see Section \ref{subsec:evolution}). A comparison of several methods was accomplished over the most currently used public data sets, taking into account the bias of the metrics used to assess face anti-spoofing performance (with several perfect results), differently from~\cite{galbally:2014} and~\cite{parveen2015face}, where the results were individually analyzed. The goal is to numerically show how far spoofing detectors got considering only face. In order to fulfill such purposes, sixty-one face anti-spoofing methods were gathered (including the works that participated in the two competitions). Previous surveys did not include any in-depth assessment of existing face spoofing detection approaches~\cite{galbally:2014,parveen2015face}, leaving unclear which ways should be followed and what need to be done in technical terms, considering only face spoofing detection. Differently from the philosophical and general discussion found in~\cite{galbally:2014}, concerning facts and challenges in the spoofing detection domain, the numerical-driven evaluation of the area allows suggesting other ways to evaluate the performance (avoiding supposedly perfect results), as well as new future research topics (\textit{e.g.}, deep learning~\cite{fan:2014}, and collaborative clustering \cite{CORNUEJOLS201881}) to be applied in face anti-spoofing methods (see Section \ref{subsec:evolution}).

\subsection{Methodology}

This compilation of works is based on a literature search in the following data sets: Scopus\footnote{http://www.scopus.com/}, IEEE Xplore\footnote{http://ieeexplore.ieee.org/Xplore/home.jsp}, Engineering Village\footnote{http://www.engineeringvillage.com/} and Google Scholar\footnote{https://scholar.google.com}. On these sources, articles were consulted considering all publications with the following keywords: \textbf{face recognition, face spoofing detection, face liveness detection, countermeasure against spoofing attacks and face anti-spoofing detection methods}. The choice of the articles was made according to the following criteria: (i) they should follow the same protocol when evaluating the study; (ii) they should indicate the results using at least one of the metrics discussed in Subsection \ref{subsec:metrics}, (iii) they should be comparable to other studies using the same data set, and finally (iv) they must be peer-reviewed.

It is noteworthy that there were two competitions on face spoofing detection referred in ~\cite{chakka:2011}, ~\cite{chingovska:2013}. The results obtained by the competition teams were analyzed, and the names of the groups and universities were used as references to the methods used in the first face spoofing detection competition, such as: Ambient Intelligence Laboratory (AMILAB), Center for Biometrics and Security Research, Institute of Automation, Chineses Academy of Sciences (CASIA), Idiap Research Institute (IDIAP), Institute of Intelligent Systems and Numerical Applications in Engineering, \textit{Universidad de Las Palmas de Gran Canaria} (SIANI), Institute of Computing, Campinas University (UNICAMP) and Machine Vision Group, University of Uolu (UOLU)~\cite{chakka:2011}. As well as, the names CASIA, Fraunhofer Institute for Computer Graphics Research (IGD), joint team from IDIAP, UOLU, UNICAMP and CPqD Telecom \& IT Solutions (MaskDown), the LNM Institute of Information Technology, Jaipur (LNMIIT), Tampere University of Technology (MUVIS), University of Cagliari (PRA Lab), \textit{Universidad Autonoma de Madrid} (ATVS) and UNICAMP refer to the teams that participated in the second face spoofing detection competition~\cite{chingovska:2013}. Throughout this text, these team names will be cited as the reference of the method in the competition (\cite{chakka:2011} or~\cite{chingovska:2013}).

\section{Face spoofing detection} \label{sec:detection}

Face spoofing detection~\cite{maatta:2011,maatta:2012,schwartz:2011,bharadwaj:2013,tirunagari:2015}, face liveness detection~\cite{yan:2012,peixoto:2011,yang:2013,wang:2013,tan:2010}, counter measure against facial spoofing attacks~\cite{komulainen:2013,pereira:2012,kose:2013,kosedugelay:2013,kosedugelay:icdsp2013}, and face anti-spoofing~\cite{chingovska:2012,erdogmus:2013,galballymarcel:2014} are all terms interchangeably used to denote methods to identify an impostor trying to masquerade him/herself as a genuine user in facial recognition systems.

\subsection{Types of face spoofing attacks}
Face spoofing systems usually consider the following types of spoofing attacks:

  \begin{itemize}

  \item The use of a {\bf flat printed photo} (see Figure~\ref{fig02}\subref{fig02b}) is the most common one, with great potential to take place, since most people have facial pictures available on the Internet ({\it e.g.,} social media) or could be photographed by an impostor without collaboration or permission. 

  \item In the \textbf{eye-cut photo} attack, eye regions of a printed photo are cut off to exhibit blink behavior of the impostor (see Figure~\ref{fig02}\subref{fig02c}).

  \item \textbf{Warped photo} attacks consist in bending a printed photo in any direction to simulate facial motion (see Figure~\ref{fig02}\subref{fig02d}).

  \item An attack via \textbf{video playback} shows almost all behaviors similar to real faces, with many of the intrinsic features of valid user movements (see Figure~\ref{fig02}\subref{fig02e}). This type of attack has physiological signs of life that are not presented in photos, such as eye blinking, facial expressions, and movements in the head and mouth, and it can be easily performed using tablets or large smartphones.
  
  \begin{figure*}[b]
  \centering{\subfloat[]{\includegraphics[height=0.15\textwidth]{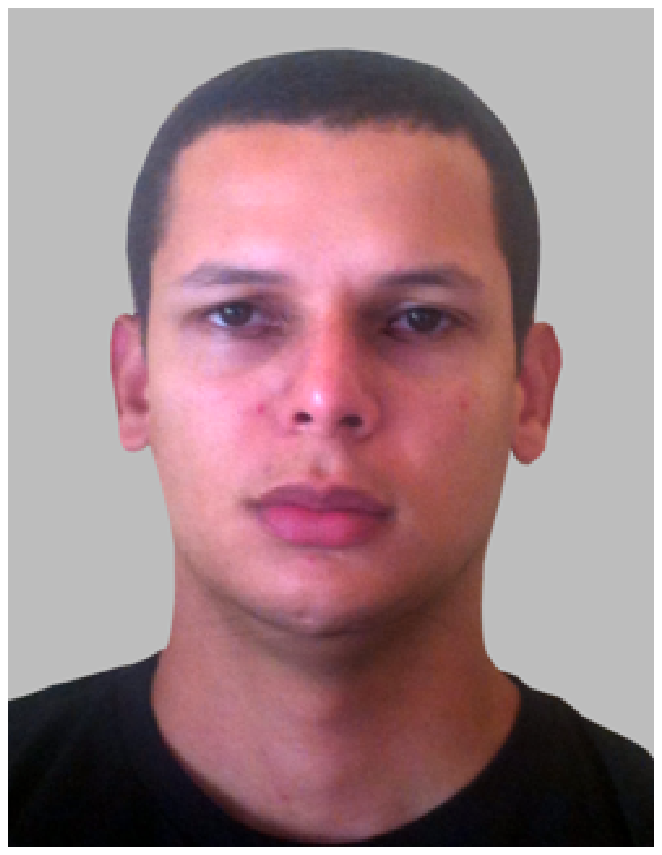} \label{fig02a}}}
  \hfil
  \centering{\subfloat[]{\includegraphics[height=0.15\textwidth]{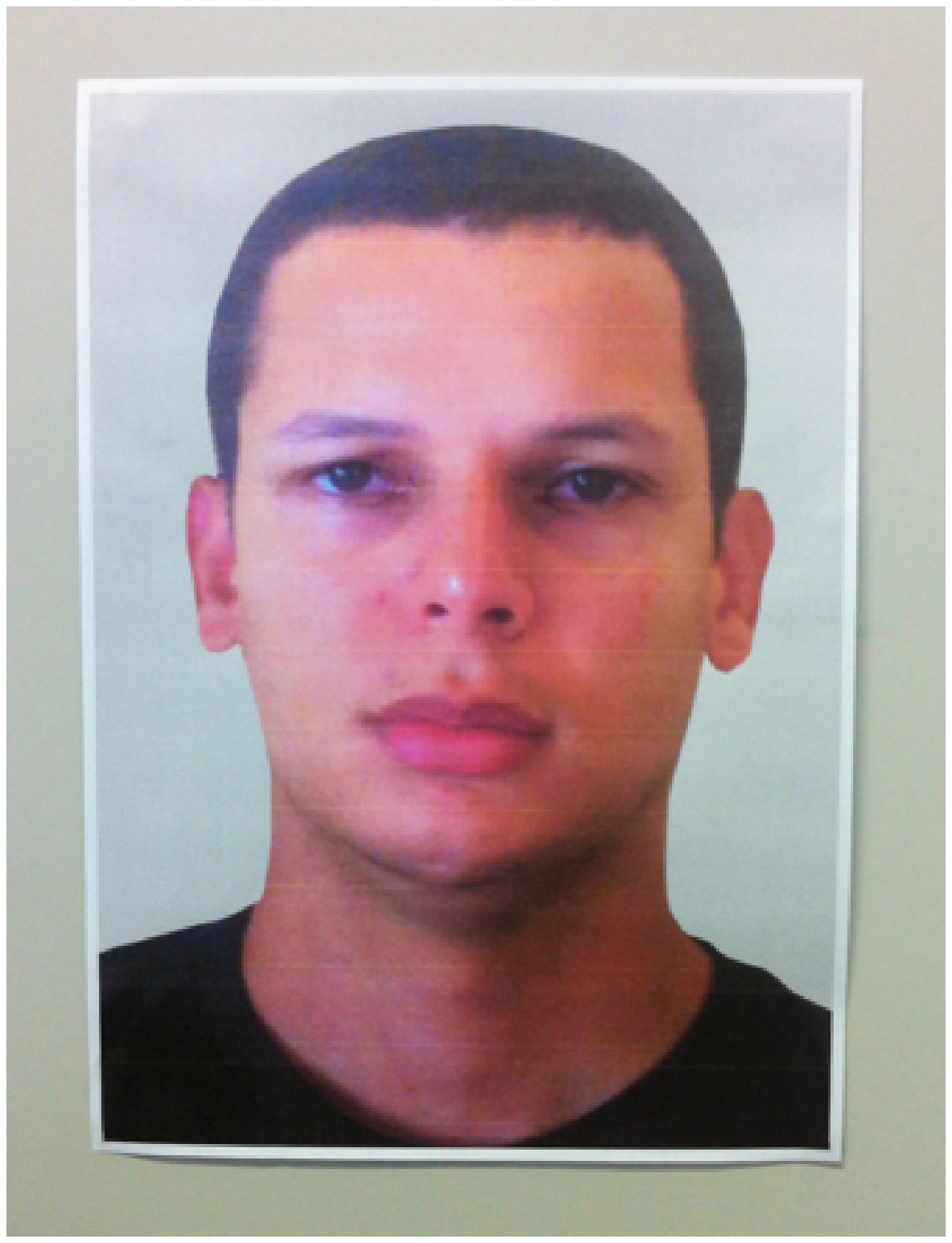} \label{fig02b}}}
  \hfil
  \centering{\subfloat[]{\includegraphics[height=0.15\textwidth]{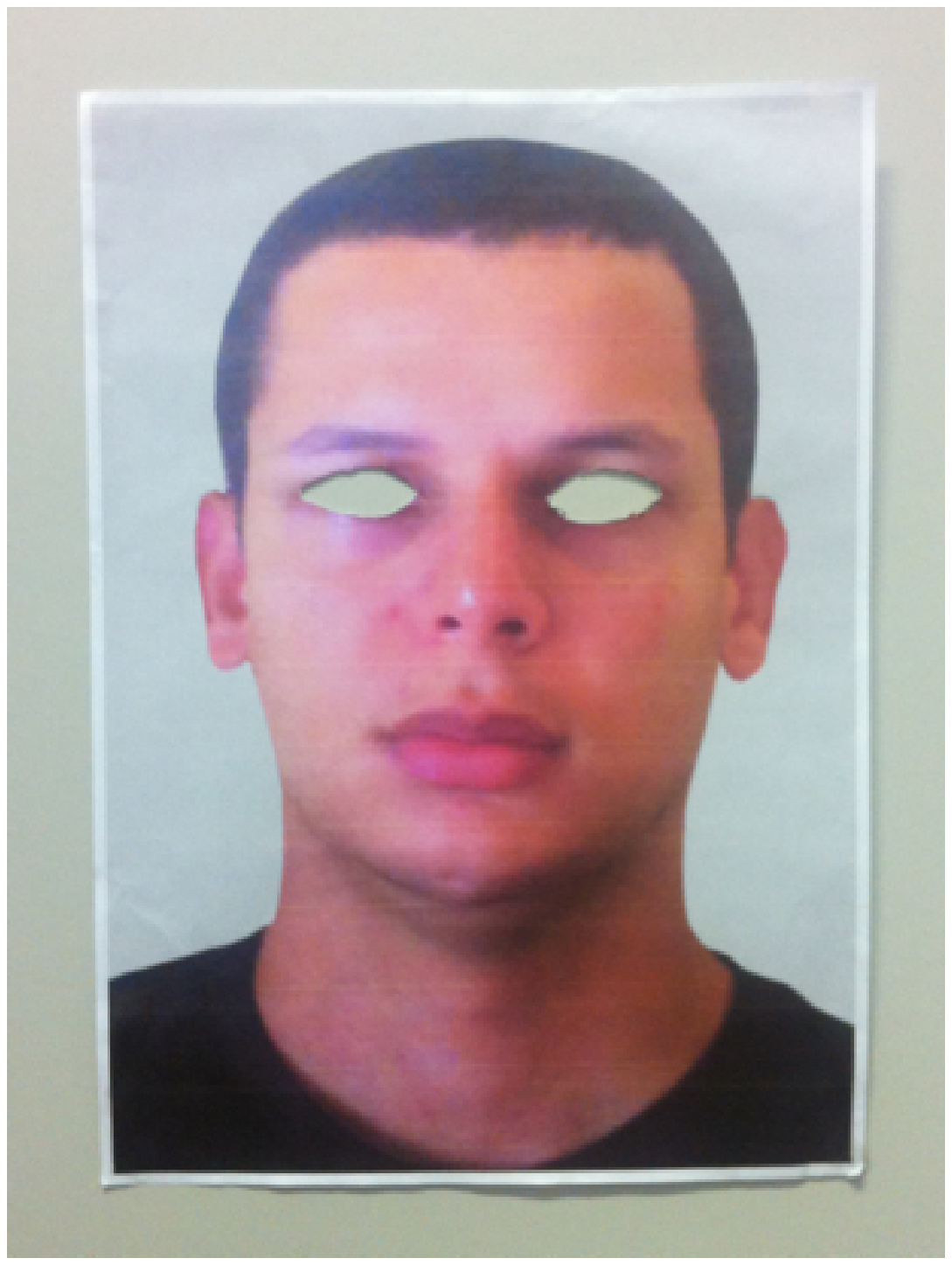} \label{fig02c}}}
  \hfil
  \centering{\subfloat[]{\includegraphics[height=0.15\textwidth]{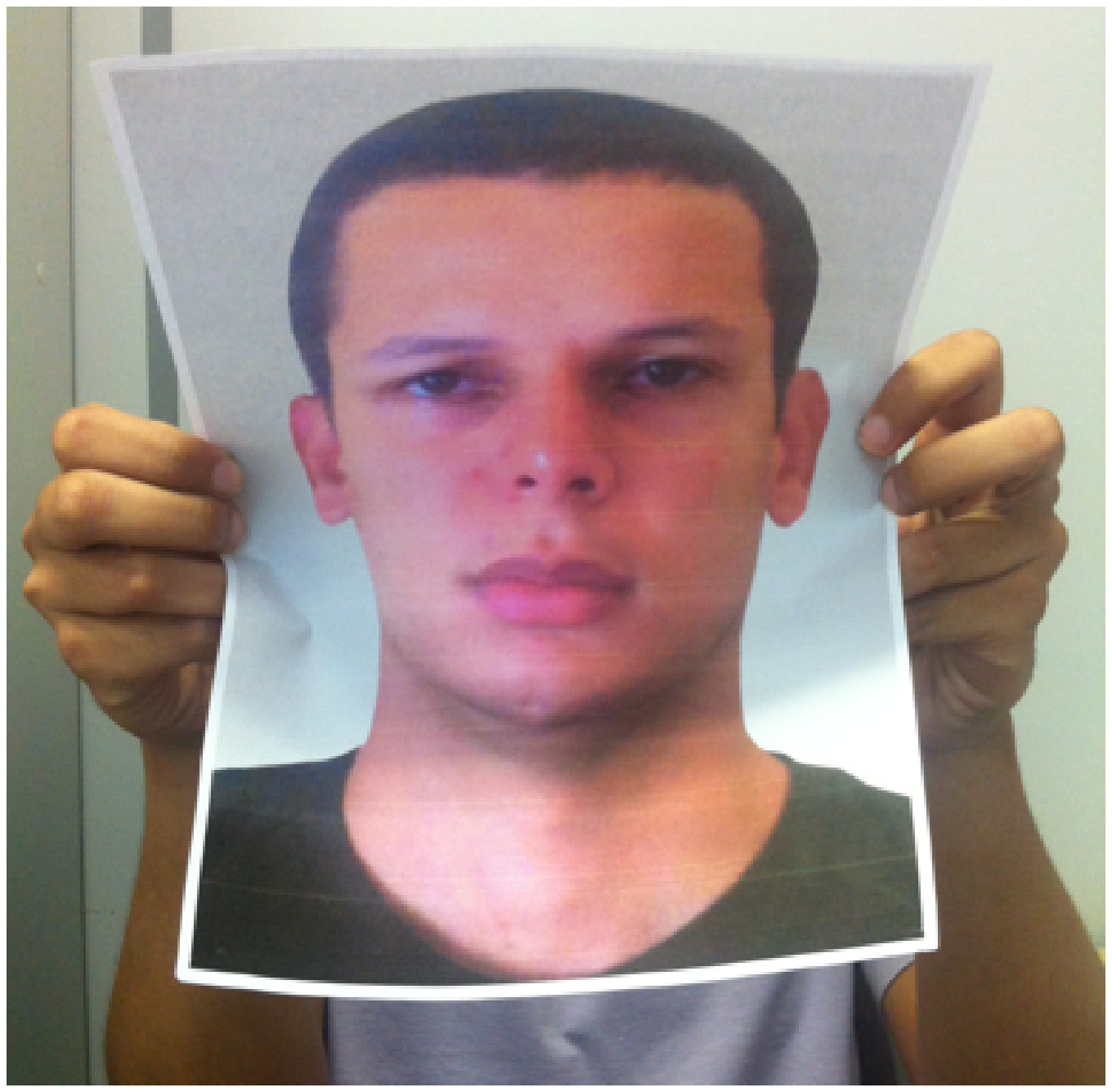} \label{fig02d}}}
  \hfil
  \centering{\subfloat[]{\includegraphics[height=0.15\textwidth]{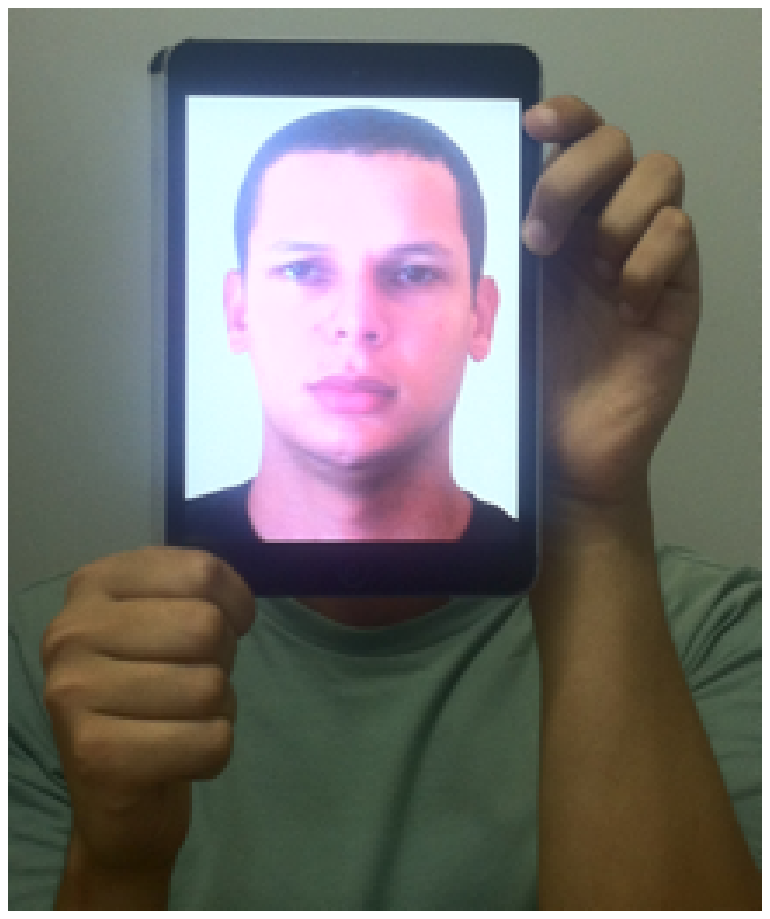} \label{fig02e}}}
  \hfil
  \centering{\subfloat[]{\includegraphics[height=0.15\textwidth]{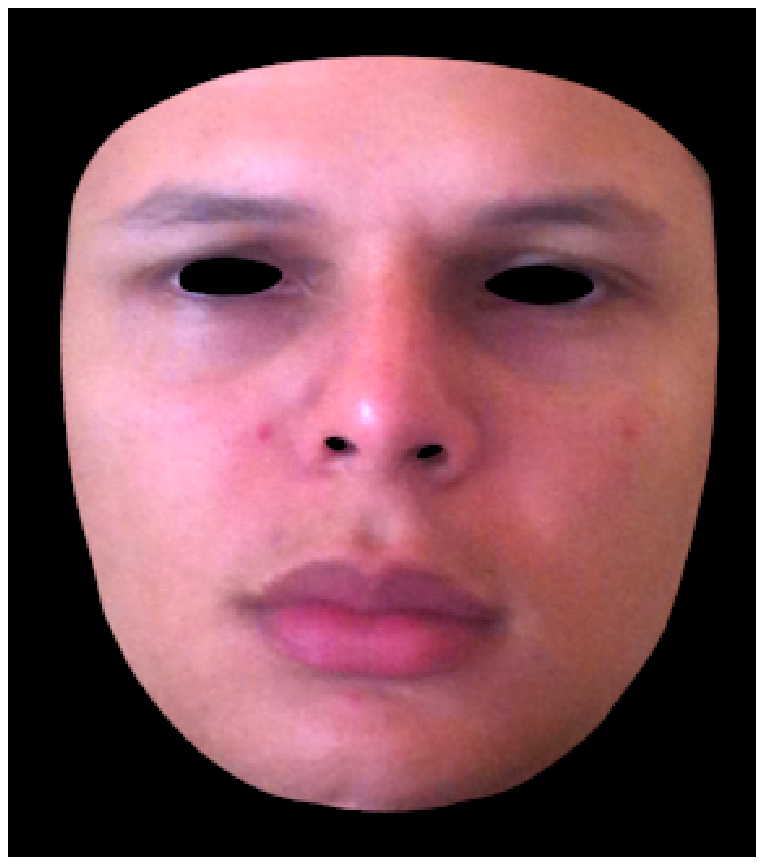} \label{fig02f}}}
  \hfil
  \centering{\subfloat[]{\includegraphics[height=0.15\textwidth]{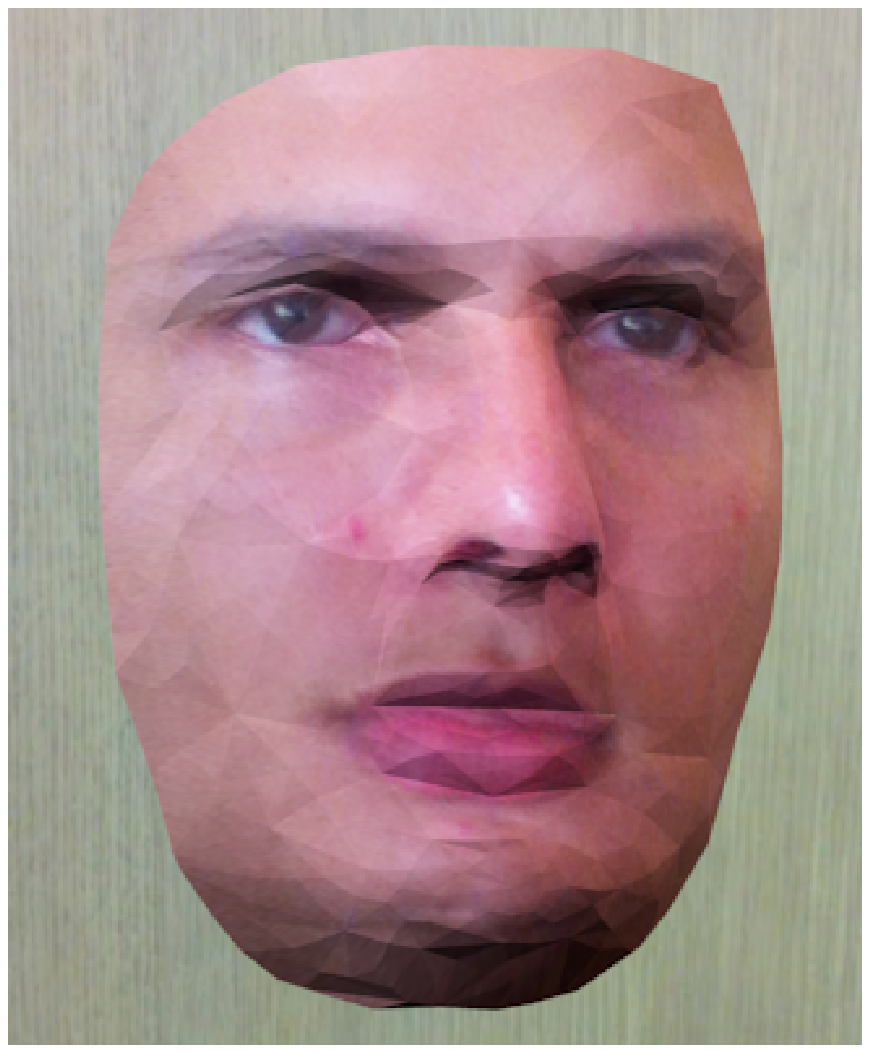} \label{fig02g}}}
  \caption{Types of face spoofing attack: (a) genuine user; (b) flat printed photo; (c) eye-cut photo; (d) warped photo; (e) video playback; (f) life-size wearable mask and (g) paper-cut mask.}
  \label{fig02}
  \end{figure*}

\begin{table*}[!tp]
  \caption{Summary of the gathered works on face spoofing detection.}
  \label{tab:01}
  \centering
 %\footnotesize
 \small
  \begin{tabular}{ll}

  \hline \\ [-2mm]
  {\bf Descriptors} & {\bf Related works} \\
  \hline \\ [-2mm]

\textbf{Texture} & 
\textbf{LBP and variations} (\cite{komulainen:2013,chingovska:2012,erdogmus:2013,kim:2012,boulkenafet2015face,souza:2017}, IDIAP ~\cite{chakka:2011}, MaskDown ~\cite{chingovska:2013}, \\&\cite{maatta:2011,maatta:2012,yang:2013,kose:2013,kosedugelay:2013,kosedugelay:2014,erdogmus2014spoofing,yang2015person,kimahn:2016}, UOULU ~\cite{chakka:2011}, CASIA~\cite{chingovska:2013}, \\& LNMIIT~\cite{chingovska:2013}, Muvis~\cite{chingovska:2013}, ~\cite{kose:2012},\cite{pereira:2012,asim:2017}, MaskDown ~\cite{chingovska:2013}), \textbf{LPQ} \cite{yang:2013,boulkenafet2016face}, \\&
\textbf{LGS}~\cite{housamlau:2014}, \textbf{ILGS}~\cite{housam:2014}, \textbf{Texture} (AMILAB ~\cite{chakka:2011},  PRA Lab ~\cite{chingovska:2013}), \\&\textbf{MLPQ-TOP}~\cite{arashloo2015face}, \textbf{MBSIF-TOP}~\cite{arashloo2015face},  \textbf{LDP-TOP}~\cite{phan:2017,phan:2016}, \\& \textbf{HOG} (\cite{maatta:2012,schwartz:2011,yang:2013,yang2015person,komulainenhadid:2013}, \textbf{Gabor Wavelets} (~\cite{maatta:2012}, Muvis ~\cite{chingovska:2013}),\\& \textbf{GLCM} (\cite{schwartz:2011,kimahn:2016,pinto:2012,pinto:2015}, UNICAMP~\cite{chakka:2011,chingovska:2013}, MaskDown~\cite{chingovska:2013}),\\&  \textbf{DoG}(~\cite{peixoto:2011,zhang:2012},UNICAMP~\cite{chakka:2011}), \textbf{HSC} (\cite{feng2016integration,schwartz:2011}, UNICAMP~\cite{chakka:2011}),\\&  \textbf{AOS}~\cite{alotaibi:2017}, \textbf{DNN}~\cite{menotti:2015}\\ [2mm]

 \textbf{Motion} & \textbf{CRF}~\cite{pan:2007}, \textbf{OFL}~\cite{feng2016integration,kollreider:2008,kollreider:2009}, \textbf{HOOF}~\cite{bharadwaj:2013}, \textbf{HMOF} (CASIA ~\cite{chingovska:2013}), \\& \textbf{RASL} (~\cite{yan:2012}, CASIA ~\cite{chakka:2011}), \textbf{Motion Correlation} (\cite{komulainen:2013}, CASIA ~\cite{chingovska:2013}),\\& \textbf{GMM} (~\cite{yan:2012,pinto2015face}, CASIA ~\cite{chakka:2011}, LNMIIT ~\cite{chingovska:2013}), \textbf{DMD}~\cite{tirunagari:2015},\\& \textbf{Motion} (SIANI ~\cite{chakka:2011}, IGD ~\cite{chingovska:2013})\\ [2mm]
   
  \textbf{Frequency} & 
  \textbf{2D-DFT} (\cite{kim:2012,phan:2017,pinto:2012,pinto:2015,pinto2015face,caetano2015face}, UNICAMP ~\cite{chingovska:2013}), \textbf{1D-FFT} (CASIA ~\cite{chingovska:2013}),\\& \textbf{2D-FFT} (LNMIIT ~\cite{chingovska:2013}), \textbf{Haar Wavelets} (~\cite{yan:2012}, CASIA ~\cite{chakka:2011}) \\ [2mm]

  \textbf{Color} & \textbf{CF}(~\cite{schwartz:2011}, UNICAMP \cite{chakka:2011}), \textbf{IDA}~\cite{kimahn:2016,wen:2015}, \textbf{IQA}~\cite{galballymarcel:2014}, \\& \textbf{IQM} (ATVS ~\cite{chingovska:2013}), \textbf{Color} (\cite{boulkenafet2015face,boulkenafet2016face,laskhminarayana:2017}, AMILAB ~\cite{chakka:2011}, PRA Lab ~\cite{chingovska:2013}) \\ [2mm]

  \textbf{Shape} & \textbf{CLM}~\cite{wang:2013}\\ [2mm]

  \textbf{Reflectance} & \textbf{Variational Retinex}~\cite{tan:2010,kosedugelay:icdsp2013,kosedugelay:2014} \\ [1mm]

  \hline  \\ [-2mm]
  {\bf Classifiers} & {\bf Related works} \\
  \hline  \\ [-2mm]

  \textbf{Discriminant} & \textbf{SVM} 
  (\cite{maatta:2012,kose:2013,kosedugelay:2013,kosedugelay:icdsp2013,kosedugelay:2014,boulkenafet2016face,komulainenhadid:2013},  CASIA ~\cite{chingovska:2013}, \cite{maatta:2011,bharadwaj:2013,pereira:2012,chingovska:2012,erdogmus:2013,kim:2012,boulkenafet2015face,pinto:2012,wen:2015}, \\& ~\cite{tirunagari:2015,phan:2017,phan:2016}, (LNMIIT ~\cite{chingovska:2013}), ~\cite{yang:2013,wang:2013,komulainen:2013,kimahn:2016,zhang:2012,pinto2015face},  UOULU~\cite{chakka:2011}, AMILAB~\cite{chakka:2011},\\& PRA Lab ~\cite{chingovska:2013}),  \textbf{LDA} (\cite{bharadwaj:2013,erdogmus:2013,galballymarcel:2014,erdogmus2014spoofing}, MaskDown~\cite{chingovska:2013}, ATVS~\cite{chingovska:2013}),  \textbf{MLP}~\cite{komulainen:2013}, \\& \textbf{NN}~\cite{feng2016integration}, \textbf{CNN}~\cite{souza:2017,asim:2017, alotaibi:2017,menotti:2015,laskhminarayana:2017}, \textbf{BN} (SIANI ~\cite{chakka:2011}), \textbf{Adaboost} (IGD ~\cite{chingovska:2013}) \\ [2mm]

  \textbf{Regression} & \textbf{LLR} ( ~\cite{komulainen:2013}, MaskDown ~\cite{chingovska:2013}), \textbf{LR} (\cite{yan:2012},  CASIA ~\cite{chakka:2011}), \textbf{SLR}~\cite{peixoto:2011}, \textbf{SLRBLR}~\cite{tan:2010},\\&  \textbf{PLS} (\cite{schwartz:2011,yang2015person,pinto:2012,pinto:2015}, UNICAMP~\cite{chakka:2011}, Muvis ~\cite{chingovska:2013}), \textbf{KDA}~\cite{arashloo2015face} \\[2mm]

  \textbf{Distance Metric} &  $\bm{\chi^2}$ (\cite{kose:2012}, IDIAP ~\cite{chakka:2011}), \textbf{Cosine}~\cite{housamlau:2014,housam:2014} \\ [2mm] 
  
  \textbf{Heuristic} & 
\textbf{Blink count}~\cite{pan:2007}, \textbf{Thresholding}~\cite{kollreider:2008,caetano2015face}, \textbf{Weighted sum}~\cite{kollreider:2009}\\ [1mm]

  \hline
  \end{tabular}
  \end{table*}
  
  \item \textbf{Mask} attacks are of two types: life-size wearable mask (see Figure~\ref{fig02}\subref{fig02f}) and paper-cut mask (see Figure~\ref{fig02}\subref{fig02g}). These attacks are addressed to anti-spoofing systems that analyze 3D face structures, being one of the most complex attacks to be detected. Mask manufacturing is much more difficult and expensive than the other types of attacks, requiring 3D scanning and printing special devices. 
  \end{itemize}
  
  \subsection{Taxonomy of face spoofing methods}
  
Face recognition systems based on 2D and 3D images can be exposed to spoofing attacks, which can be verified by different approaches. In order to summarize them, we organized all gathered works in terms of descriptors and classifiers. Descriptors were categorized as texture, motion, frequency, color, shape or reflectance; while classifiers are organized as discriminant, regression, distance metric or heuristic. Table~\ref{tab:01} presents the summary of the proposed taxonomy, and the descriptors and classifiers are respectively discussed and analyzed in Sections~\ref{subsec:descriptors} and~\ref{subsec:classification}.

  \subsection{Descriptors}
  \label{subsec:descriptors}

  \subsubsection{Texture}
 
  Texture features are extracted from face images under the assumption that printed faces produce certain texture patterns that do not exist in real ones. Texture is probably the strongest evidence of spoofing, since more than 69\% of the works (see Table~\ref{tab:01}) use texture alone or combine it with other descriptors in their countermeasures.

Different texture descriptors can be used to detect facial spoofing, but local binary patterns (LBP) \cite{ojala1996acomparative} is the very first choice, as it can be observed in Table \ref{tab:01}. Indeed, nearly half of the surveyed works explore LBP or any of its variations. LBP is a grayscale, illumination-invariant, texture-coding technique that labels every pixel by comparing it with its neighbors, concatenating the result into a binary number. The number of neighbors, neighborhood radius, and coding strategy are all parameters of the method. The final computed labels are then organized in histograms to describe the texture, which can be performed for the entire image or even image paths. Different LBP configurations can be found in spoofing detection, such as the original LBP (\cite{komulainen:2013,chingovska:2012,erdogmus:2013,kim:2012,boulkenafet2015face,souza:2017}, IDIAP ~\cite{chakka:2011}, MaskDown ~\cite{chingovska:2013}), multi-scale LBP (\cite{maatta:2011,maatta:2012,yang:2013,kose:2013,kosedugelay:2013,kosedugelay:2014,erdogmus2014spoofing,yang2015person,kimahn:2016}, UOULU ~\cite{chakka:2011}, CASIA~\cite{chingovska:2013}, LNMIIT~\cite{chingovska:2013}, Muvis~\cite{chingovska:2013}), LBP variance (LBPV) ~\cite{kose:2012}, and LBP from three orthogonal planes (LBP-TOP) (\cite{pereira:2012,asim:2017}, MaskDown ~\cite{chingovska:2013}). LBP-TOP can be considered a hybrid texture-motion descriptor, since it combines both spatial and temporal information. Other texture-coding techniques were also explored for spoofing detection, including local phase quantization (LPQ)~\cite{yang:2013,boulkenafet2016face}, which uses invariant blurring properties when extracting features from images. This descriptor is a phase information of the locally calculated Fourier spectrum for each position of the pixel in the image. Different approaches include local graph structure (LGS)~\cite{housamlau:2014} and improved LGS (ILGS)~\cite{housam:2014}x', which were used to extract texture features by comparing a target pixel and its neighboring pixels. Other methods have also used texture descriptors in face spoofing detection competitions to analyze face spoofing attacks (AMILAB ~\cite{chakka:2011}, PRA Lab ~\cite{chingovska:2013}). Multiscale local phase quantization on three orthogonal planes (MLPQ-TOP) is an extension of LPQ for time-varying texture analysis, which explores the blur-insensitive characteristic of the Fourier phase spectrum \cite{arashloo2015face}. Multiscale binarised statistical image feature descriptor on three orthogonal planes (MBSIF-TOP) use filters based on statistical learning that represent spatio-temporal texture descriptors \cite{arashloo2015face}. Dynamic texture descriptor can be found in local derivative pattern on three orthogonal
planes (LDP-TOP) \cite{phan:2017,phan:2016}. LDP-TOP analyzes discriminative textures of spectrum videos, where subtle face movements occur over frames.

Histograms of oriented gradients (HOG)~\cite{maatta:2012,schwartz:2011,yang:2013,yang2015person,komulainenhadid:2013} is another texture descriptor that represents the variation of gradient orientations in different parts of the image in an illumination-invariant fashion. As such, the magnitude of the gradients in different orientations are summed in cells, which are lately combined in blocks. Bins, cells and blocks are normalized at the end to compose the final feature vector. 

Gabor wavelets have been also applied in multiple scales and orientations in order to extract texture information in image cells. Usually, Gabor wavelets are calculated using mean and standard deviation of the magnitude of the coefficients at multiple scales and orientation (\cite{maatta:2012}, Muvis ~\cite{chingovska:2013}).

A compact and discriminant global representation can be achieved in the gray level co-occurrence matrices (GLCM) ~\cite{haralick1973texture}. GLCM describes the joint probability of neighboring pixels, and different Haralick features can be extracted from each matrix (\cite{schwartz:2011,kimahn:2016,pinto:2012,pinto:2015}, MaskDown~\cite{chingovska:2013}, UNICAMP ~\cite{chingovska:2013}). 

Edge information can also be considered for texture representation. In order to describe edges, difference of gaussians (DoG) are used to remove lighting variations while preserving high frequency components~\cite{peixoto:2011,zhang:2012}, and histograms of shearlet coefficients (HSC) are used to estimate the distribution of edge orientations in a multi-scale analysis~\cite{feng2016integration,schwartz:2011}. Nonlinear diffusion based on additive operator
splitting (AOS) \cite{alotaibi:2017} is also used to extract edge information for spoofing detection, applying a large time step to speed up the diffusing process and to distinguish the edges and surface texture in the input image.

Finally, following a very recent trend in computer vision field, deep neural networks (DNN)~\cite{menotti:2015} are trained in order to provide adaptive features which describe trainable texture.
  
 \subsubsection{Motion}

  Table~\ref{tab:01} clearly shows that motion descriptors are the second in importance for face spoofing detection, and there are two different ways of considering motion for this purpose. One way is to detect and describe intra-face variations, such as eye blinking, facial expressions and head rotation. Conditional random fields (CRF) have been recently used to determine eye closity and consequently detect blinking~\cite{pan:2007}; for global facial movements, optical flow of lines (OFL) is used to measure spatio-temporal variations of face images in horizontal and vertical orientations~\cite{feng2016integration,kollreider:2008,kollreider:2009}, histogram of oriented optical flow (HOOF) and histogram of magnitudes of optical flows (HMOF) are applied to create a binned representation of facial motion directions and magnitudes (CASIA~\cite{chingovska:2013}); and robust alignment by sparse and low-rank decomposition (RASL) tries to align faces in multiple frames and measure non-rigid motion (\cite{yan:2012}, CASIA ~\cite{chakka:2011}). 
  
  Another way of using motion is to evaluate the consistency of the user interaction within the environment. In light of that, motion correlation between face and background regions is computed (\cite{komulainen:2013}, CASIA ~\cite{chingovska:2013}), as well as, traditional background subtraction based on gaussian mixture models (GMM) (\cite{yan:2012, pinto2015face},  CASIA ~\cite{chakka:2011}, LNMIIT ~\cite{chingovska:2013}). 
 
Facial texture of an individual within a sequence of  frames is explored by using the dynamic mode decomposition (DMD) \cite{tirunagari:2015}, which extracts features by means of eigenfaces in the snapshots displaced on temporal-spatial. DMD is used in combination with LBP technique as a texture descriptor, which is applied to capture evidences of human presence in a video sequence, such as eye blink and movements of the lips. Finally, in competitions, other types of methods have been used as a motion descriptor to analyze face spoofing attacks (SIANI ~\cite{chakka:2011}, IGD ~\cite{chingovska:2013}).

  \subsubsection{Frequency}

  Frequency-based countermeasures take advantage of certain image artifacts that occur in spoofing attacks. 2D discrete Fourier transform (2D-DFT), and 1D and 2D fast Fourier transform (1D-FFT, 2D-FFT) are calculated to find these artifacts in single ~\cite{kim:2012} or multiple images (\cite{phan:2017,pinto:2012,pinto:2015,pinto2015face,caetano2015face}, CASIA~\cite{chingovska:2013}, LNMIIT~\cite{chingovska:2013}, UNICAMP~\cite{chingovska:2013}). When one considers multiple images, the concept of Visual Rhythms is used to merge multiple Fourier spectra in a single map that represents spatial frequency information over time, and then HOG, LBP and/or GLCM can be used for final face representation. When specifically considering color banding, which concerns abrupt changes caused by inaccurate print or screen flicker, Haar wavelets decomposition can be applied to find large unidirectional variations (\cite{yan:2012}, CASIA ~\cite{chakka:2011}).

  \subsubsection{Color}

  Although colors do not remain constant due to lighting variations, certain dominant characteristics are considerable clues to discriminate impostors from genuine faces. In this context, color frequency (CF) histograms describe the distribution of colors in an image(~\cite{schwartz:2011}, UNICAMP~\cite{chakka:2011}). These histograms are computed for different blocks of the image, as performed by HOG, using three bins to encode the number of pixels with the highest gradient magnitude in each color channel. Image moments globally describe face liveness by means of image distortion analysis (IDA)~\cite{kimahn:2016,wen:2015}, image quality assessment (IQA)~\cite{galballymarcel:2014} and image quality measures (IQM) (ATVS ~\cite{chingovska:2013}). IDA was proposed to extract characteristics through the HSV and RGB color spaces, smoothing and light intensity. IQA allows to maximize both critical performance measures in a complete face spoofing detection. IQM aims to show that the lowest values, obtained by quality measurements, produced with Gaussian filtering are samples of impostor face. YCbCr and HSV color spaces are used as color descriptors in \cite{boulkenafet2015face,boulkenafet2016face}. In \cite{laskhminarayana:2017}, each channel of RGB color space was used for feature extraction. Other methods have been used as color descriptors, used in competitions, to analyze face spoofing attacks (AMILAB ~\cite{chakka:2011}, PRA Lab ~\cite{chingovska:2013}).

  \subsubsection{Shape}

  Shape information is very useful to deal with printed photo attacks, since facial geometry can not be reproduced in a planar surface. Active contours based on constrained local models (CLM) are used to detect facial landmarks in a video sequence. These landmarks define then a sparse 3D structure that describes the planarity of the face~\cite{wang:2013}.

  \subsubsection{Reflectance}

  Considering that genuine and impostor faces behave differently in the same illumination conditions, it is possible to use the reflectance information to distinguish them. To accomplish that, the Variational Retinex method decomposes an input image into reflectance and illumination components~\cite{tan:2010,kosedugelay:icdsp2013,kosedugelay:2014} in order to analyze the entire image.

  \subsection{Classifiers} \label{subsec:classification}

  \subsubsection{Discriminant}
  
  The idea behind discriminant techniques is to distinguish different classes by minimizing intra-class variation and/or maximizing inter-class variation. This type of classifier is explored in approximately 64\% of the gathered works. 
  
  As evidenced in Table~\ref{tab:01}, works use a discriminant classifier alone or along with others in their frameworks. Support vector machines (SVM) are the most common classification technique in spoofing detection, and often presents superior performance. In order to achieve that, SVM finds optimal hyperplanes to separate descriptors from genuine and impostor faces. When these classes are not linearly separable, different kernel functions can be used to obtain a nonlinear classifier. Although linear SVM has been extensively used in different countermeasures (\cite{maatta:2012,kose:2013,kosedugelay:2013,kosedugelay:icdsp2013,kosedugelay:2014,boulkenafet2016face,komulainenhadid:2013,pinto:2012}, CASIA ~\cite{chingovska:2013}), radial basis function kernel (\cite{maatta:2011,bharadwaj:2013,pereira:2012,chingovska:2012,erdogmus:2013,kim:2012,boulkenafet2015face,pinto:2012,wen:2015}), and histogram intersection kernel~\cite{tirunagari:2015,phan:2017,phan:2016} have also been applied to increase the classification accuracy. Different SVM versions can also be considered, such as Hidden Markov Support Vector Machines (LNMIIT ~\cite{chingovska:2013}). In (~\cite{yang:2013,wang:2013,komulainen:2013,kimahn:2016,zhang:2012,pinto2015face},  UOULU~\cite{chakka:2011}, AMILAB~\cite{chakka:2011}, PRA Lab ~\cite{chingovska:2013} ), however, the authors do not describe the type of SVM kernel used in the experiments.
  
  As an alternative to linear approaches, the linear discriminant analysis (LDA) (\cite{bharadwaj:2013,erdogmus:2013,galballymarcel:2014,erdogmus2014spoofing}, MaskDown~\cite{chingovska:2013}, ATVS ~\cite{chingovska:2013}) explicitly models the difference between classes and within classes to address the classification task, with an advantage of being used for dimensionality reduction. 

Other types of classifiers use discriminant procedures to accomplish face spoofing detection: multilayer perceptron (MLP)~\cite{komulainen:2013} was used to evaluate whether excessive movement (flat printed photo-strike by hand) or no movement (flat printed photo strike attached to a media) had variations during an $N$ video sequence; neural network (NN)~\cite{feng2016integration} is good at learning implicit patterns, which is able to recognize motion cues for spoofing detection with proper training. NN is trained by a backpropagation procedure using a labeled data set through an autoencoder, which is treated as a pre-training process; convolutional neural networks (CNN)~\cite{souza:2017,asim:2017,alotaibi:2017, menotti:2015,laskhminarayana:2017} uses trainable features with shared weights and local connections between different layers, where all weights in all layers of a CNN network are learned through training. A CNN aims to learn invariance representations of scale, translation, rotation and related transformations on a trainable-based feature framework. Bayesian network (BN) (SIANI ~\cite{chakka:2011}) provides a probabilistic method by extending Bayes’s rule for updating probabilities in the light of new evidences. Adaboost is a type of ensemble classifier that speeds up the process of finding discrimination of impostor and genuine users (IGD ~\cite{chingovska:2013}).

  \subsubsection{Regression}

The regression-based classification maps use input descriptors directly into their class labels considering a predictive model obtained from known pairs of descriptors and labels. They have been widely used for spoofing detection due to their simplicity, accuracy and efficiency. Different regression methods are referred in the literature: linear logistic regression (LLR) (\cite{komulainen:2013}, MaskDown ~\cite{chingovska:2013}) was used for the combination of information extracted through two descriptors; for the combination of correlative motion and texture (\textit{e.g.}, LBP) applications, using MLP and SVM classifiers, respectively; logistic regression (LR) (\cite{yan:2012}, CASIA ~\cite{chakka:2011}) is a confidence quantification of every feature representation, which in the final scores are fused by weight sum rule, and the weights of different classifiers are learned on validation set using grid search; sparse logistic regression (SLR)~\cite{peixoto:2011} analyzes different lighting conditions and regions of high frequency for detecting images made by impostors; sparse low rank bilinear logistic regression (SLRBLR)~\cite{tan:2010} was explored in images with prominent reflectance and illumination, using two techniques to extract these characteristics of the image, being reflectance based on variational retinex, and the illumination based on the DoG technique in the identification of medium-high frequency bands; partial least square (PLS) (\cite{schwartz:2011,yang2015person,pinto:2012,pinto:2015}, UNICAMP ~\cite{chakka:2011}, Muvis ~\cite{chingovska:2013}) is calculated from a linear transformation on the features extracted by descriptors using weighting methods. 

Kernel discriminant analysis (KDA) is a regression classifier, which projects the input data onto a discriminative spectral subspace, avoiding the computational time found in eigen-analysis \cite{arashloo2015face}. In other words, KDA uses projective functions (vectors) based on eigen-decomposition of kernel matrix, being costly when applied to a large number of training samples.

  \subsubsection{Distance Metric}

 The use of distance metrics is supposed to improve the performance in face spoofing detection systems, with the goal of measuring the dissimilarities among samples. However, these approaches usually require an exhaustive search to accomplish the classification task, which may lead to a high cost in large reference sets. Chi-square ($\chi^2$) (\cite{kose:2012}, IDIAP ~\cite{chakka:2011}) and cosine distance~\cite{housamlau:2014,housam:2014} are common choices to this end, and they are used to compute the cumulative distance of a probe face (that one to be identified) and the entire reference set to decide whether the face is genuine or impostor.  

  \subsubsection{Heuristic}

  Different heuristics have been used to decide whether a face is real or fake. As a drawback, heuristics may lead to overfitting, specially when only self-collected data is considered. Number of eye blinks~\cite{pan:2007}, motion measurements thresholding~\cite{kollreider:2008}, average pixel ratio thresholding \cite{caetano2015face} and weighted sum of motion measurements~\cite{kollreider:2009} are examples of heuristics found in the literature.

\section{Quantitative evaluation of the surveyed works} \label{sec:comparative}

  Results reported in the surveyed papers were grouped according to the data sets used in their experiments. All numerical values in our study are the exact same values presented in their original works, which followed the same evaluation protocol.

  \subsection{Data sets}\label{subsec:datasets}

  Nine publicly available data sets were chosen to evaluate the methods: Concerning 2D attacks, NUAA Photograph Imposter~\cite{tan:2010}, Yale Recaptured~\cite{peixoto:2011}, Print-Attack~\cite{anjos:2011}, Replay-Attack~\cite{chingovska:2012}, Casia Face Anti-Spoofing~\cite{zhang:2012}, MSU-MFSD~\cite{wen:2015} and UVAD (\cite{pinto:2015},\cite{pinto2015uvad1}) are the most known and used in the literature; with respect to mask attacks, Kose and Dugelay's data set~\cite{kose:2013} and 3D Mask Attack data set~\cite{erdogmus:2013} are the only two found in the literature. General characteristics of each data set are summarized in Table \ref{tab2}, and more details can be found in ~\cite{peixoto:2011,tan:2010,kose:2013,chingovska:2012,erdogmus:2013,zhang:2012,anjos:2011}. 

  \begin{table*}[t]
  \caption{Summary of available face spoofing data sets.}
  \label{tab2} 
  \centering
  %\scriptsize
  \footnotesize
  \begin{tabular}{|c|l|c|c|l|}
     \hline 
   \textbf{Year} & \textbf{Data set} & \multicolumn{1}{c|}{\textbf{\#Subjects}}& \multicolumn{1}{c|}{\textbf{\#Real/Fake}} & \multicolumn{1}{c|}{\textbf{Type of attack}} \\
  \hline 
  \centering

  2010 & NUAA Photograph Imposter ~\cite{tan:2010} & 15 & 5105/7509 & 1. Flat printed photo \\ & & & & 2. Warped photo \\ \hline

  2011 & Yale Recaptured~\cite{peixoto:2011} & 10  & 640/1920 & 1. Flat printed photo\\ \hline

  2011 & Print-Attack ~\cite{anjos:2011} & 50 & 200/200 & 1. Flat printed photo  \\ \hline

  2012 & Replay-Attack~\cite{chingovska:2012} & 50 &200/1000 &1. Flat printed photo \\ & & & & 2. Video playback \\ \hline

  2012 & Casia Face Anti-Spoofing ~\cite{zhang:2012} &50 & 150/450  & 1. Warped photo  \\ & & & &2. Eye-cut photo \\ & & & & 3. Video playback \\ \hline

  2013 & Kose and Dugelay ~\cite{kose:2013}  & 20 & 200/198 & 1. Mask \\ \hline

  2013 & 3D Mask Attack ~\cite{erdogmus:2013} & 17 & 170/85 & 1. Mask \\ \hline
  
  2014 & MSU-MFSD ~\cite{wen:2015} & 35& 70/210 &1. Flat printed photo \\ & & & & 2. Video playback \\ \hline   
  
  2015 & UVAD \cite{pinto:2015}, \cite{pinto2015uvad1} & 404 & 808/16268 & 1. Video playback  \\ \hline
  
  \end{tabular}
  \end{table*}

  NUAA Photograph \textit{Imposter} data set\footnote{$http://parnec.nuaa.edu.cn/xtan/NUAAImposterDB\_download.html$}~\cite{tan:2010} is one of the first publicly available data sets for face spoofing detection evaluation. Images in NUAA were collected by cheap webcams in three sessions on different environments and under different illumination conditions, with an interval of two weeks between each session. The evaluated attack is a printed photo, which can be flat or warped. These photo attacks were prepared using A4 paper and a color printer.

The main goal of the Yale Recaptured data set\footnote{$http://ic.unicamp.br/~rocha/pub/downloads/2011\-icip/$}~\cite{peixoto:2011} was to have impostor images in multiple illumination conditions. As such, texture-based methods are commonly employed over this data set. Static images were collected with a distance of 50 centimeters between the display and the camera.

  The Print-Attack data set\footnote{$https://www.idiap.ch/dataset/printattack/download\-proc$} \cite{anjos:2011} was used to benchmark different works in the first spoofing detection competition \cite{chakka:2011}. This data set was created by showing a flat printed photo of a genuine user to an acquisition sensor in two ways: Hand-held ({\it i.e.,} the impostor holds the photo using the hands) or fixed support ({\it i.e.,} photos are stuck on a wall).

  Replay-Attack data set\footnote{$https://www.idiap.ch/dataset/replayattack/download\-proc$}~\cite{chingovska:2012} is an extension of the Print-Attack data set to evaluate spoofing in videos and photos, and it was used in the second spoofing detection competition~\cite{chingovska:2013}. It consists of 1,300 video clips of photo and video attacks. All images and videos were collected under different lighting conditions, and three different attacks modes were considered: printed photo in high-resolution and video playbacks, using a mobile phone with low-resolution screen, and a 1024$\times$768 pixels ipod screen.

  Casia Face Anti-Spoofing data set\footnote{$http://www.cbsr.ia.ac.cn/english/FaceAntiSpoofdata sets.asp$}~\cite{zhang:2012} contains seven scenarios with different types of attack and a variety of image qualities. This data set presents three types of attacks: warped photo, eye-cut photo and video playback. Kose and Dugelay's data set\footnote{$http://www.morpho.com/$}~\cite{kose:2013} is a paid-mask data set created by the MORPHO company. Subjects were captured by a 3D scanner that uses a structured light technology to obtain genuine images of facial shape and texture. After that, masks for those images were manufactured by Sculpteo 3D Printing\footnote{$http://www.sculpteo.com/en/$}, and then recaptured by the same sensor to obtain impostor images.

  3D Mask-Attack data set (3DMAD)\footnote{$https://www.idiap.ch/dataset/3dmad/download-proc$}~\cite{erdogmus:2013} was the first publicly available data set for mask attacks, and it consists of video sequences recorded by an RGB-D camera. Masks were manufactured using the services of ThatsMyFace\footnote{$http://www.thatsmyface.com/Products/products.html$}, and one frontal and two profile images of each subject for that purpose were required.
  
  MSU Mobile Face Spoofing data set (MSU MFSD)\footnote{$http://www.cse.msu.edu/rgroups/biometrics/Publications/data sets/MSUMobileFaceSpoofing/index.htm$}\cite{wen:2015} consists of 280 video clips of print photo and video attack attempts to 35 participants. All printed photos used for attacks were created with a state-of-the-art color printer on larger sized paper. In order to perform an attack, video playback from each participant was taken under the similar conditions as in their authentication sessions.
  
  Unicamp Video-Attack data set (UVAD)\footnote{$http://figshare.com/articles/visualrhythm_antispoofing/1295453$}\cite{pinto:2015,pinto2015uvad1} is comprised of videos of valid accesses and attacks of 404 subjects, all built at Full HD quality, recorded at 30 frames per second and nine seconds long. All videos were created by
filming each person in two sections under different lighting conditions, backgrounds and places (indoors and outdoors).

\begin{table*}[!th]
\caption{Summary of non-public face spoofing data sets.}
\label{tab20} 
%\centering
%\scriptsize
\footnotesize
\begin{tabular}{|c|l|c|c|l|}
\hline

 \textbf{Year} & \textbf{data set} & \multicolumn{1}{|c|}{\textbf{\#Subjects}}& \multicolumn{1}{|c|}{\textbf{\#Real/Fake}}   & \textbf{Types of attacks} \\
 
\hline
\centering

2012 & Pinto et al. ~\cite{pinto:2012} & 50  & 100/600  & 1. Video playback  \\ \hline

2012 & BERC Webcam~\cite{kim:2012} & 25 & 1408/7461 & 1. Flat printed photo \\ \hline

2012 & BERC ATM~\cite{kim:2012} & 20 & 1797/5802 & 1. Flat printed photo \\ \hline

2013 & Wang et al. ~\cite{wang:2013}  & 50  & 750/2250  & 1. Flat printed photo  \\ & & &  & 2. Warped photo \\ \hline

\end{tabular}
\end{table*}

Other non-public data sets are proposed in \cite{wang:2013,kim:2012,pinto:2012}. In ~\cite{pinto:2012}, the data set was introduced to detect video-based spoofing. Kim et al. ~\cite{kim:2012} proposed two different data sets called BERC Webcam and BERC ATM, which consist of images taken from live people and four types of 2-D paper masks (photo, print, magazine, caricature). Wang et al. ~\cite{wang:2013} created an image data set for photo attack evaluation using flat-printed and warped photos. Table \ref{tab20} summarizes the main characteristics of the aforementioned data sets. 

  \subsection{Performance Metrics}\label{subsec:metrics}

\begin{figure}[t]
  \centering
  \includegraphics[width=5.0cm]{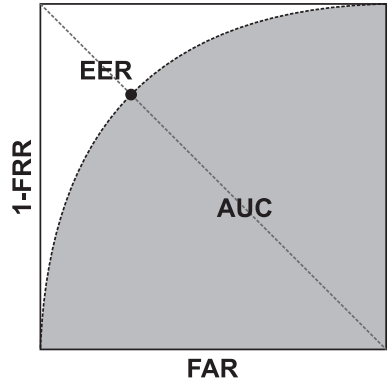} 
  \caption{Relation among the metrics on the ROC curve.}
  \label{figroc}
  \end{figure}

  A spoofing detection system is subject to two types of errors: an impostor can be accepted as a genuine user ({\it i.e.} number of false acceptance - NFA), or a genuine user can be considered as an impostor ({\it i.e.,} number of false rejection - NFR). The probability of these errors to occur are respectively called false acceptance rate (FAR) and false rejection rate (FRR). These rates present an inversely proportional relation. A receiver operating characteristics (ROC) curve is obtained by computing all possible pairs of FAR and FRR values, as illustrated in Fig.~\ref{figroc}. The integral of a ROC curve is known as the area under curve (AUC), {\it i.e.,} the gray-filled area in Fig.~\ref{figroc}. Also, the point of the ROC curve where FAR equals FRR is called equal error rate (EER), and the point where the average of FAR and FRR is minimal is called half total error rate (HTER). Finally, the overall accuracy (ACC) considers both genuine users and impostors along with the FAR and FRR. Table \ref{tab3} summarizes all the aforementioned metrics.

 \begin{savenotes}
\begin{table*}[t]
   \caption{Metrics commonly applied on face spoofing evaluation.}
   \label{tab3} 
   \centering
   %\tiny
   
   \footnotesize
   \begin{tabular}{|l|c|c|c|}
   \hline 
    \textbf{Metric} & \multicolumn{1}{c|}{\textbf{Stand for}}& \multicolumn{1}{c|}{\textbf{Equation}} &  \multicolumn{1}{c|}{\textbf{Type}}  \\
   \hline
   \centering 
   FAR	& False Acceptance Rate	& $FAR=\dfrac{NFA}{\#Impostor}$ & Error \\ \hline

   FRR	& False Rejection Rate	& $FRR=\dfrac{NFR}{\#Genuine)}$ & Error \\ \hline

   EER	& Equal Error Rate & $EER=(FAR=FRR)$& Error\\ \hline

   HTER &	Half Total Error Rate &	$HTER=\dfrac{FAR + FRR}{2}$ & Error \\ \hline

   ACC & Accuracy & $100 \times \Bigg( 1 - \frac{FAR \times \#Impostor + FRR \times \#Genuine}{\#Impostor + \#Genuine} \Bigg)$ & Hit \\ \hline

   AUC	& Area Under Curve	& $Area=\int_{a}^{b} f(x) \hspace{1mm}dx$, where $f$ : [$a$,$b$] $\rightarrow$ $\mathbb{R}$ & Hit 
   \\ \hline
   \end{tabular}
   \end{table*}
   \end{savenotes}
 
  Since most of the considered data sets are not balanced ({\it i.e.}, the number of impostors and genuine images is different), ACC may lead to a biased performance analysis. All other metrics are based on a separate evaluation of FAR and FRR, so they are more reliable for a comparative analysis. For these reasons, surveyed works were compared using metrics according to the following order of preference: EER, HTER, AUC and ACC.

  \subsection{Performance analysis of the existing spoofing detectors}\label{subsec:benchmarks}

  Comparing different works is a difficult task, since most of the time we do not have access to the original source codes, and reproducing codes and experimental results are very complicated. For that reason, we have decided to perform this comparison by using the results reported in the gathered papers. However, determination of the best method based on the reported results is not an easy task. It is possible to make mistakes even when comparing works that use the very same data set, specially if this data set is prone to be biased~\cite{torralba:2011}. Strictly speaking, besides a common available data set, it is of underlying importance to follow the same methodology and to have the same metrics when comparing different countermeasures. 

Given the data sets presented in Section~\ref{subsec:datasets}, some criteria were adopted to select which works should be considered in our analysis: (i) they must follow the same data set protocol; (ii) they must report its results using at least one of the metrics discussed in Section~\ref{subsec:metrics}; and (iii)they  must be comparable to other works using the same data set ({\it i.e.} use the most common metrics for that data set). On that account, some works were then removed from the analysis for NUAA~\cite{chingovska:2012,housamlau:2014,housam:2014}, Print-Attack~\cite{yan:2012,yang:2013}, Replay-Attack \cite{yang2015person,caetano2015face}, Casia Face Anti-Spoofing~\cite{tirunagari:2015,chingovska:2012,galballymarcel:2014,yang2015person,pinto2015face,wen:2015,laskhminarayana:2017}, Kose and Dugelay's~\cite{kosedugelay:2013}, 3DMAD ~\cite{erdogmus2014spoofing,pinto2015face} data sets and other works in \cite{pan:2007,kollreider:2008, kollreider:2009}. Therefore, such works are not presented in Table \ref{tab:01} and Fig. \ref{fig03}. Tables~\ref{tab4}-\ref{tab14} summarize the selected results. It is noteworthy that sometimes it was necessary to take conclusions by indirectly comparing different metrics, as in Table~\ref{tab4}.

 %---------------------------NUAA-----------------
  \begin{savenotes}
  \begin{table*}[!th]
    \centering
    \footnotesize
    \caption{Results over NUAA Imposter data set.}
     \label{tab4}%

      \begin{tabular}{lll|ccc}

  \hline \\ [-2.8mm]
      \textbf{Reference} & \textbf{Features} & \textbf{Classifier} & \textbf{EER (\%)}  & \textbf{AUC} & \textbf{ACC (\%)} \\ \hline

  2010, Tan {\it et al.}~\cite{tan:2010}				& Variational Retinex & SLRBLR & - & 0.94 & - \\ \hline
  2011, Maatta {\it et al.}~\cite{maatta:2011}		& LBP & SVM & 2.90 & 0.99 & 98.00 \\ \hline     
  2011, Peixoto {\it et al.}~\cite{peixoto:2011}		& DoG & SLR & - & - & 93.20 \\ \hline
  2011, Schwartz {\it et al.}~\cite{schwartz:2011}	& CF + HOG + HSC + GLCM & PLS & 8.20 & 0.96 & - \\ \hline
\textbf{2012, Maatta \textbf{\textit{et al.}} ~\cite{maatta:2012}}						& \textbf{LBP + Gabor Wavelets + HOG} & \textbf{SVM} & \textbf{1.10} & \textbf{0.99} & \textbf{-} \\ \hline
  2012, Kose and Dugelay~\cite{kose:2012}				& LBPV & $\chi$ $^{2}$ & 11.97 & - & - \\ \hline
  2013, Yang {\it et al.}~\cite{yang:2013}			& LBP + LPQ + HOG & SVM & 1.90 & 0.99 & 97.70 \\ \hline
  2015, Arashloo {\it et al.}~\cite{arashloo2015face}	& MLPQ-TOP + MBSIF-TOP & KDA & 1.80 & - & - \\ \hline
  2017, Alotaibi and&&&&\\ Mahmood~\cite{alotaibi:2017}	& AOS & CNN & - & - & 99.00 \\ \hline
  2017, Souza {\it et al.}~\cite{souza:2017}	& LBP & CNN	 & 1.80 & 0.99 & 98.20 \\ \hline  
  
  \end{tabular}
  \end{table*}
  \end{savenotes}
   %---------------------------YALE---------------------------
  \begin{table*}[t]
    \centering
     \footnotesize
        \caption{Results over Yale Recaptured data set.}
     \label{tab5}%
      \begin{tabular}{lll|c}
  \hline \\ [-2.8mm]
      \textbf{Reference} & \textbf{Features}& \textbf{Classifier} & \textbf{ACC (\%)} \\ \hline
  2011, Peixoto {\it et al.}~\cite{peixoto:2011}	& DoG & SLR & 91.70 \\ \hline     
\textbf{2012, Maatta \textbf{\textit{et al.}}~\cite{maatta:2012}}	& \textbf{LBP + Gabor Wavelets + HOG} & \textbf{SVM} & \textbf{100.00} \\ \hline
  \end{tabular}
  \end{table*}
%------------------
  %---------------------------Print-Attack----------------------
  \begin{savenotes}
  \begin{table*}[t]
    \centering
     \footnotesize
    \caption{Results over Print-Attack data set.}
     \label{tab6}%

      \begin{tabular}{lll|c}
  \hline \\ [-2.8mm]
      \textbf{Reference} & \textbf{Features}& \textbf{Classifier} & \textbf{HTER (\%)} \\ \hline

\textbf{2011,  IDIAP~\cite{chakka:2011}}						& \textbf{LBP} & \textbf{$\chi$ $^{2}$} & \textbf{0.00} \\ \hline

\textbf{2011,  UOULU~\cite{chakka:2011}}						& \textbf{LBP} & \textbf{SVM}  &  \textbf{0.00} \\ \hline

\textbf{2011,  CASIA~\cite{chakka:2011}}						& \textbf{RASL + GMM + } & \textbf{LR} & \textbf{0.00} \\& \textbf{Haar Wavelets} &  &  \\ \hline

  2011,  AMILAB~\cite{chakka:2011}					& Color + Texture & SVM & 0.63 \\ \hline
  2011,  SIANI~\cite{chakka:2011}						& Motion & BN & 10.63 \\ \hline
  2011,  UNICAMP~\cite{chakka:2011} and&&&\\ Schwartz et al.~\cite{schwartz:2011}	& CF + HOG + HSC + GLCM & PLS  & 0.63 \\ \hline

\textbf{2012, Maatta \textbf{{\textit et al.}}~\cite{maatta:2012}}			& \textbf{LBP + Gabor Wavelets + HOG} & \textbf{SVM} & \textbf{0.00} \\ \hline
  2013, Bharadwaj {\it et al.}~\cite{bharadwaj:2013}		& HOOF & LDA &  0.62 \\ \hline
  \textbf{2015, Tirunagari \textbf{{\textit et al.}}~\cite{tirunagari:2015}}	& \textbf{DMD + LBP} & \textbf{SVM} & \textbf{0.00} \\ \hline
  \end{tabular}
  \end{table*}
  \end{savenotes}
  
Table~\ref{tab4} summarizes the results of the methods concerning the NUAA data set, and the most common metrics were EER, AUC and ACC. This data set presents a relatively balanced number of positives and negatives samples, which avoids biased results when using ACC. Peixoto {\it et al.}~\cite{peixoto:2011} did not report both EER and AUC, but their ACC shows that they did not achieve the best performance. As it can be observed in Table~\ref{tab4}, methods with high AUC have low EER. Although AUC does not allow us to differ between the methods proposed by Maatta {\it et al.}~\cite{maatta:2011,maatta:2012} and Yang {\it et al.}~\cite{yang:2013}, EER clearly shows that Maatta {\it et al.}~\cite{maatta:2012} achieved the best performance for the NUAA data set.
  
  A summary of the results considering the Yale Recaptured data set is presented in Table \ref{tab5}. Works using this data set were compared by means of ACC, the only metric in common to all of them. Since this data set is highly unbalanced ({\it i.e.,} a ratio of 1:3), ACC would not be the most recommended metric. For this comparison, however, the use of ACC is not an issue due to the perfect performance reported ~\cite{maatta:2012}, which means that both classes were perfectly classified.

 %---------------------------Replay-Attack---------------------------
  \begin{savenotes}
  \begin{table*}[tb]
    \centering
  \footnotesize
    \caption{Results over Replay-Attack data set.}
     \label{tab7}

      \begin{tabular}{lll|c}
  \hline \\ [-2.8mm]

  \textbf{Reference} & \textbf{Features}& \textbf{Classifier} & \textbf{HTER (\%)} \\ \hline 

  2012, Chingovska {\it et al.}~\cite{chingovska:2012}	& LBP & SVM & 15.16 \\ \hline

  2012, Freitas {\it et al.}~\cite{pereira:2012}			& LBP-TOP & SVM & 7.60 \\ \hline

  2013, Komulainen {\it et al.}~\cite{komulainen:2013}	& Motion Correlation + LBP & LLR + SVM & 5.11 \\&& + MLP &  \\ \hline

  2013, Bharadwaj {\it et al.}~\cite{bharadwaj:2013}		& HOOF + LBP & LDA & 1.25 \\ \hline

\textbf{2013,  CASIA~\cite{chingovska:2013}}					& \textbf{LBP + 1D-FFT + HMOF} & \textbf{SVM} &\textbf{0.00} \\& \textbf{ + Motion Correlation} &  &  \\ \hline

  2013,  IGD~\cite{chingovska:2013}					& Motion & Adaboost   & 9.13 \\ \hline

  2013,  MaskDown~\cite{chingovska:2013}				& LBP + GLCM + LBP-TOP & LLR + LDA & 2.50   \\ \hline

\textbf{2013,  LNMIIT ~\cite{chingovska:2013}	}			& \textbf{LBP + GMM + 2D-FFT} & \textbf{SVM} & \textbf{0.00} \\ \hline

  2013,  Muvis~\cite{chingovska:2013}					& LBP + Gabor Wavelets & PLS & 1.25 \\ \hline

  2013,  PRA Lab~\cite{chingovska:2013}				& Color + Texture & SVM & 1.25 \\ \hline

  2013,  ATVS~\cite{chingovska:2013}					& IQM & LDA & 12.00 \\ \hline

  2013,  UNICAMP~\cite{chingovska:2013}				& 2D-DFT + GLCM & SVM & 15.62 \\ \hline

  2014, Galbally {\it et al.}~\cite{galballymarcel:2014}	& IQA & LDA & 15.20 \\ \hline

  2015, Menotti {\it et al.}~\cite{menotti:2015}			& DNN & CNN  & 0.75 \\ \hline

  2015, Tirunagari {\it et al.}~\cite{tirunagari:2015}	& DMD + LBP & SVM & 3.75 \\ \hline

  2015, Wen {\it et al.}~\cite{wen:2015}					& IDA & SVM & 7.41 \\ \hline

  2015, Pinto {\it et al.}~\cite{pinto:2015}				& 2D-DFT & PLS & 14.27 \\ \hline
  
  2015, Boulkenafet {\it et al.}~\cite{boulkenafet2015face}	& LBP + Color & SVM & 2.90\\ \hline
  
  2015, Arashloo {\it et al.}~\cite{arashloo2015face}	& MLPQ-TOP + MBSIF-TOP & KDA & 1.00\\ \hline
  2015, Pinto {\it et al.}~\cite{pinto2015face}	& GMM + 2D-DFT & SVM & 2.75 \\ \hline

2016, Boulkenafet {\it et al.}~\cite{boulkenafet2016face}	& LPQ + Color & SVM & 3.30\\ \hline

\textbf{2016, Feng \textbf{{\textit et al.}}~\cite{feng2016integration}}	& \textbf{HSC + Optical Flow}  & \textbf{NN} & \textbf{0.00}\\ \hline

2016, Kim {\it et al.} ~\cite{kimahn:2016}	& MLBP + GLCM + IDA & SVM & 5.50 \\\hline

2016, Phan {\it et al.} ~\cite{phan:2016}	& LDP-TOP & SVM & 1.75 \\\hline

2017, Alotaibi and&&&\\ Mahmood~\cite{alotaibi:2017}	& AOS & CNN & 10.00 \\\hline 
2017, Lakshminarayana{\it et al.} ~\cite{laskhminarayana:2017}	& Color & CNN & 0.80 \\\hline

  \end{tabular}
  \end{table*}
  \end{savenotes}

 %---------------------------Casia---------------------------
  \begin{savenotes}
  \begin{table*}[t]
    \centering
    %\scriptsize

    \caption{Results over Casia Face Anti-Spoofing data set.}
     \label{tab8}

      \begin{tabular}{lll|c}
  \hline \\ [-2.8mm]
      \textbf{Reference} & \textbf{Features}& \textbf{Classifier} & \textbf{EER (\%)} \\ \hline

2012, Zhang {\textit et al.}~\cite{zhang:2012}               & DoG & SVM      & 17.00   \\ \hline

  2013, Komulainen {\it et al.}~\cite{komulainenhadid:2013} & HOG  & SVM & 3.30     \\ \hline

  2013, Yang {\it et al.}~\cite{yang:2013}                  & LPQ + LBP + HOG  & SVM & 11.80 \\ \hline
  
   2015, Boulkenafet {\it et al.}~\cite{boulkenafet2015face}	& LBP + Color & SVM & 6.20\\ \hline

\textbf{2016, Boulkenafet \textbf{\textit {et al.}}~\cite{boulkenafet2016face}}	& \textbf{LPQ + Color} & \textbf{SVM} & \textbf{3.20}\\ \hline

2016, Feng \textit{et al.}~\cite{feng2016integration}	& HSC + Optical Flow & NN & 5.83 \\ \hline

2016, Kim {\it et al.} ~\cite{kimahn:2016}	& MLBP + GLCM + IDA & SVM & 4.89 \\\hline

2016, Phan {\it et al.} ~\cite{phan:2016}	& LDP-TOP & SVM & 8.94 \\\hline

2017, Asim \textit{et al.}~\cite{asim:2017} & LBP-TOP & CNN & 8.02 \\ \hline

  \end{tabular}
  \end{table*}
  \end{savenotes}
  
  %---------------------Kose and Dugelay's---------------------------
  \begin{savenotes}
  \begin{table*}[!tp]
    \centering
     \footnotesize
       \caption{Results of different methods over Kose and Dugelay's data set.}
     \label{tab9}%

      \begin{tabular}{lll|cc}
  \hline \\ [-2.8mm]
      \textbf{Reference} & \textbf{Features}& \textbf{Classifier}  & \textbf{AUC} & \textbf{ACC (\%)} \\ 
            \hline

  2013, Kose and Dugelay~\cite{kosedugelay:icdsp2013}	& Variational Retinex & SVM & 0.97 & 94.47 \\ \hline
  2013, Kose and Dugelay~\cite{kose:2013}				& LBP & SVM & 0.98 & 93.50 \\ \hline 
\textbf{2014, Kose and Dugelay~\cite{kosedugelay:2014}}		& \textbf{LBP +}&\textbf{SVM} & \textbf{0.99} & \textbf{98.99}\\& \textbf{Variational Retinex} & & &  \\ 

      \hline
      \end{tabular}%
    \label{tab:addlabel}%

  \end{table*}
  \end{savenotes}
  %-------------------------3D Mask Attack---------------------------
  \begin{table*}[th]
    \centering
     \footnotesize
        \caption{Results over 3D Mask Attack data set.}
    \label{tab10}
      \begin{tabular}{lcc|c}
  \hline \\ [-2.8mm]
       \textbf{Reference} & \textbf{Features} &  \textbf{Classifier} & \textbf{HTER (\%)} \\ \hline

  2013, Erdogmus and Marcel~\cite{erdogmus:2013}	& LBP & LDA & 0.95 \\ \hline
\textbf{2015, Menotti \textbf{\textit{et al.}}~\cite{menotti:2015}}	& \textbf{DNN} & \textbf{CNN} & \textbf{0.00} \\ \hline
\textbf{2016, Feng \textbf{\textit{et al.}}~\cite{feng2016integration}}	& \textbf{HSC + Optical Flow} & \textbf{NN} & \textbf{0.00} \\ \hline
      \end{tabular}%

  \end{table*}

  %-------------------------MSU and UVAD Attack---------------------------
  \begin{table*}[t]
 
    \centering
     \footnotesize
        \caption{Results over UVAD and MSU-MFSD data set.}
    \label{tab14}
      \begin{tabular}{llcc|cc}
  \hline \\ [-2.8mm]
       \textbf{Data set} &\textbf{Reference} & \textbf{Features} &  \textbf{Classifier} & \textbf{HTER (\%)}&\textbf{EER (\%)} \\ \hline
 
 UVAD & 2015,&GMM&&&\\& Pinto \textit{et al.}~\cite{pinto2015face}&  + 2D-DFT & SVM &29.87& - \\  \hline

 UVAD & 2017,&2D-DFT&&&\\& Phan \textit{et al.}~\cite{phan:2017}&  + LDP-TOP & SVM &23.69& - \\  \hline

MSU- & 2015,&&&&\\MFSD& Wen \textit{et al.}~\cite{wen:2015} & IDA& SVM & - & 5.82 \\
\hline

MSU-& 2016,&LPQ&&&\\MFSD& Boulkenafet \textit{et al.}~\cite{boulkenafet2016face}	&  + Color & SVM & - & 3.50\\ \hline

MSU-& 2016,&&&&\\ MFSD& Phan {\it et al.} ~\cite{phan:2016}	&  LDP-TOP & SVM & 7.70 & 6.54\\ \hline

\textbf{MSU-}& \textbf{2016},&\textbf{MLBP}&&&\\ \textbf{MFSD}& \textbf{Kim \textbf{\textit{et al.}}~\cite{kimahn:2016}}	& \textbf{GLCM + IDA} & \textbf{SVM} & - & \textbf{2.44}\\ \hline

      \end{tabular}%

  \end{table*}
%------------------------------- SELF COLLECTED
  \begin{table*}[t]

    \centering
       \caption{Results of different methods over other non-public data sets.}
    \label{tab15}
      \begin{tabular}{lcc|cc}
  \hline \\ [-2.8mm]
       \textbf{Data set} & \textbf{Features} &  \textbf{Classifier} & \textbf{EER 			(\%)}&\textbf{ACC (\%)} \\ \hline

    BERC Webcam ~\cite{kim:2012} & LBP + 2D DFT  & SVM  &   8.43  &   - \\ \hline
    
    BERC ATM ~\cite{kim:2012} & LBP + 2D DFT & SVM &   4.42  &  -    \\ 	\hline
    
	\textbf{Self Collected ~\cite{pinto:2012}} & \textbf{GLCM + 2D DFT}  & \textbf{PLS}  & -& \textbf{100.00}   \\ \hline
    
	\textbf{Self Collected ~\cite{pinto:2012}} & \textbf{GLCM + 2D DFT} & \textbf{SVM}   & -& \textbf{100.00}    \\  \hline

	\textbf{Self Collected ~\cite{wang:2013}} & \textbf{CLM} &  \textbf{SVM}  & -& \textbf{100.00}    \\ \hline

      \end{tabular}%

  \end{table*}
  
%------------------------------------
 As stated in Section~\ref{subsec:datasets}, the Print-Attack data set was used as a benchmark in the first spoofing detection competition~\cite{chakka:2011}, wherein three works achieved perfect score ({\it i.e.,}  IDIAP~\cite{chakka:2011},  UOULU~\cite{chakka:2011} and  CASIA~\cite{chakka:2011}). Later works~\cite{maatta:2012,tirunagari:2015} also achieved 0\% of HTER (see Table~\ref{tab6}), and Maatta {\it et al.}~\cite{maatta:2012} reached the best performance in a third data set. As shown in Table~\ref{tab2}, NUAA, Yale Recaptured and Print-Attack data sets are solely based on printed photo attacks. Given the work in~\cite{maatta:2012} achieved the lowest error rates in all of the attacks using the same approach, it is safe to assume that multiple texture features ({\it i.e.,} LBP, Gabor wavelets and HOG) and an SVM classifier are enough to detect printed photo attacks over that data set. 
 
 The Replay-Attack data set was used in the second spoofing detection competition~\cite{chingovska:2013}, and both CASIA and LNMIIT obtained 0\% of HTER. There is a proposed method in ~\cite{feng2016integration}, which also achieved a perfect HTER (see Table~\ref{tab7}). This data set has an uneven number of real and fake images ({\it i.e.,} a ratio of 1:5), but it does not influence the analysis since all works report their results  using the same metric.
 
  The Casia Face Anti-Spoofing data set is characterized by the highest number of types of attacks, as presented in Table~\ref{tab2}, but presenting a low number of samples. Table~\ref{tab8} presents the results on this data set, and Boulkenafet {\it et al.}~\cite{boulkenafet2016face} proposed the method with the best performance, reaching perfect results ({\it i.e.,} 3.20\% EER).
  
  Tables~\ref{tab9} and~\ref{tab10} present, respectively, the spoofing detection results for mask attacks over Kose and Dugelay's and 3D Mask Attack data sets. The best performance over that data set was achieved by the method based on texture and reflectance descriptors, and an SVM classifier. The number of fake images in the 3D Mask Attack data set was greater than the number of real ones, but works were evaluated using HTER; so unbalancing is not a problem. Menotti {\it et al.}~\cite{menotti:2015} obtained the best performance by combining deep learning and SVM. Methods dealing with video playback and mask attacks did not rely only on texture descriptors, exploring different features ({\it i.e.,} motion, frequency and reflectance) to reduce the classification error. SVM is still the most preferred classifier.
  
Table \ref{tab14} summarizes the results over UVAD and MSU-MFSD datasets. The first one was only used by one work, while in the second Kim \textit{et al.}~\cite{boulkenafet2016face} achieved the lowest EER. Table \ref{tab15} shows results over non-public data sets, and it is only presented for completeness, since it is not easy reproducible.

\section{Discussion and analysis}
\label{subsec:evolution}

Most of the effort to address the problem of face spoofing detection have been carried out over the past decade. Henceforth we provide a big picture of the field (trends), as well as the analysis of open issues and future perspectives that could be tackled and followed in order to leverage the face spoofing systems.

\subsection{Timeline and trends of the state-of-the-art works}

\begin{figure*}[ht]
\centering
\includegraphics[width=1.25\textwidth]{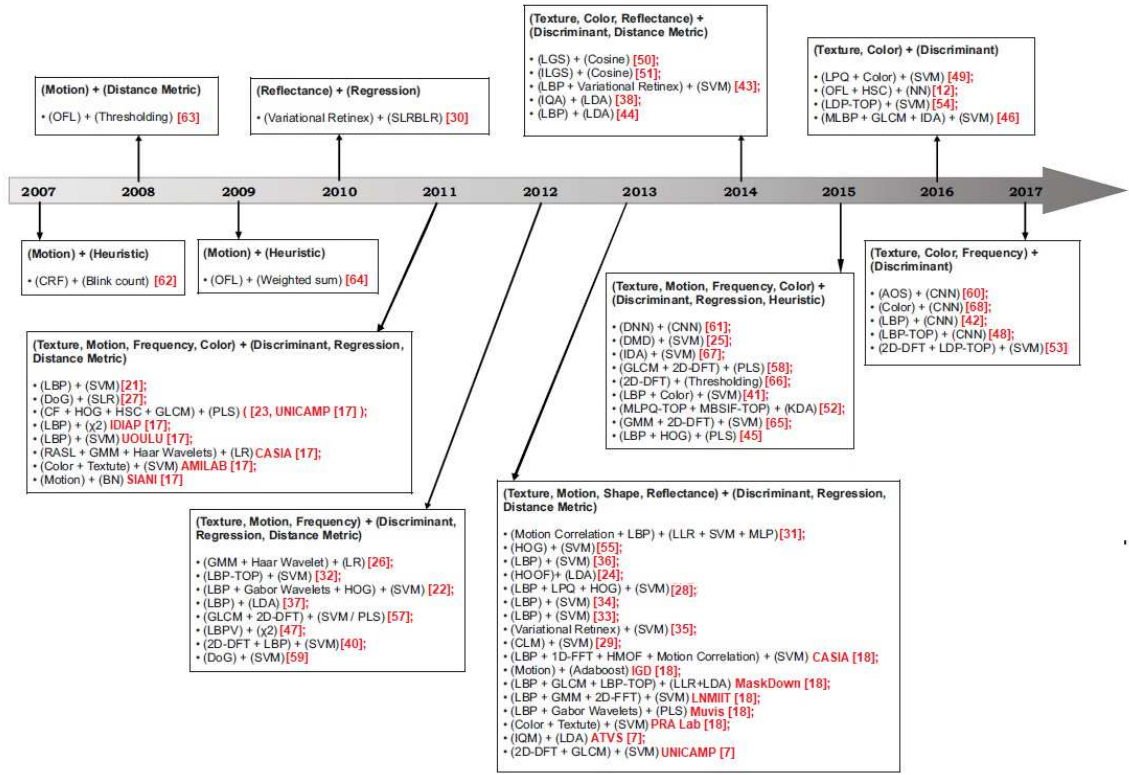}
\caption{Timeline of face spoofing detection in the last decade.}
\label{fig03}
\end{figure*}

Figure \ref{fig03} depicts a chronological arrangement of the surveyed works in order to demonstrate the convergence of descriptors and classifiers over the time.

From 2007 to 2010, spoofing detectors were mostly focused on the analysis of motion or reflectance, since both types of descriptors are based on a quite straightforward observation: printed faces do not behave or reflect light as real faces do. Although such countermeasures have persisted to date, another image cue has grown in importance in the literature: face texture. As pointed out by Tan {\it et al.}~\cite{tan:2010}, an impostor face is captured by a camera twice, while a genuine-user once. The former consequently produces artifacts that are not presented in real face acquisition. These artifacts are very perceptible in texture images, and texture coding techniques seem to be an effective way to capture and describe them, as evidenced in the number of works relying on traditional texture description approaches and their variations from 2011 to 2017.

In terms of classification, SVM-based works became more and more popular to the point of dominating the face spoofing literature in recent years, which is somehow expected, since SVM has gained a wide attention in many other machine learning tasks, such as medical diagnosis~\cite{sweilam:2010}, object recognition~\cite{muralidharan:2011} and market analysis~\cite{huang:2005}. In fact, even if we consider only face processing applications, there are several ways of exploring SVM: face recognition~\cite{tefas:2001}, face detection~\cite{osuna:1997}, facial landmark extraction~\cite{rapp:2011}, facial expression analysis~\cite{kotsia:2007}, and so forth. Although SVM provides very accurate results, the two most researched and up-to-date of these applications ({\it i.e.,} detection and recognition) are recently getting better results using deep learning methods~\cite{zhang:2014,taigman:2011}. Thus, we expect to see an attention shift towards deep learning in spoofing detection works for the next few years, which can already be seen in the most recent literature~\cite{menotti:2015,alotaibi:2017,laskhminarayana:2017,souza:2017,asim:2017}.

\subsection{Open issues}
 
After surveying all existing face spoofing detection methods in the last decade, it is still difficult to establish if there was a remarkable progress in this field. The main points that support this view are the following:

\begin{enumerate}
\item Automatic detection of impostors by face still follows the same recipe as many other computer vision problems: first extracting some features for further classifying them by a supervised predictor. Moreover, most works follow the same architecture of popular face recognition systems, using similar feature sets and classification methods. This is even more evident if one observes Table~\ref{tab11}, where the best performing works over all data sets considered in our review are shown. As stated in Section~\ref{subsec:evolution}, texture-based descriptors and SVM-based classification have prevailed in the face spoofing literature. The combination texture+SVM has reached the best performance in five out of the nine data sets analyzed. For the remaining two, texture and SVM are still there, but combined with other descriptors ({\it i.e.}, motion, frequency or reflectance).

\item Most of the time, spoofing detection follows face recognition trends. For instance, deep learning techniques are becoming very popular in face recognition~\cite{zhi-peng:2014} and have consistently outperformed other existing methods, just like what happened to LBP+SVM few years before. As demonstrated, Menotti {\it et al.}~\cite{menotti:2015} recently employed deep learning for face spoofing detection, evaluating the performance of the proposed method on two data sets: Replay-Attack and 3D Mask Attack. Their results were comparable to the state-of-the-art in the first one, and are the best performance so far in the second one. This practically shows that any face recognition breakthroughs will lead to improvements on texture-based spoofing detection as well.

\item All surveyed works perform training and testing using the same data set (although in a non-overlapping way). They presented their results with different metrics ({\it i.e.} ACC, AUC, HTER and EER) and near perfect results were found for each of the nine publicly available data sets considered in this work. Far from showing that face spoofing detection is a solved problem, this fact actually indicates the lack of a challenging data set that allows a thoroughly analysis of the proposed methods. Other computer vision problems have been conducted in this direction, like person re-identification with VIPER data set~\cite{ma:2015} and object recognition with Caltech-256 Object Category data set~\cite{griffin:2007}, both with state-of-the-art accuracy below 50\%. We believe that a large data set in a wild scenario is more likely to promote breakthroughs. In addition to a large amount of images and/or videos, multiple types of attacks should be covered, be diverse in terms of ethnicity, age and gender, and present real-world scenarios with different environments, acquisition devices, lighting conditions, and human behaviors.

\item A lack of a standard evaluation protocol for spoofing detection methods is also an issue. Currently, most of the researchers use HTER and EER for detection results to avoid biased results when a data set is unbalanced, but these metrics do not show the effects of spoofing detection on the recognition step. Chingovska {\it et al.}~\cite{chingovska:2006} introduced an evaluation protocol for biometric systems under spoofing attacks that simultaneously analyzes both recognition and spoofing detection results through expected performance and spoofability curves (EPSC) by dividing a data set in three categories: genuine users, zero-effort impostors and spoofing attacks. However, the proposed evaluation method depends on a prior probability of the spoofing attacks, or a cost relation between the ratio of incorrectly accepted zero-effort impostors and the ratio of incorrectly accepted spoofing attacks. The latter ones could vary for different systems, adding more variables to the problem. Hence, a more intuitive and self-explanatory evaluation metric is also required to instigate future efforts in this research topic. This also extends to benchmarks that rigorously evaluate both recognition and spoofing detection, which are currently not available in the literature.
\end{enumerate}

\begin{table*}[t]
\centering
 \footnotesize
\caption{Best performing works over different data sets}
\label{tab11}%
\begin{tabular}{llll}
\hline \\ [-2.8mm]
\textbf{Reference} & \textbf{Features}& \textbf{Classifier} & \textbf{Data set} \\ \hline
Maatta {\it et al.}~\cite{maatta:2012} & LBP + 	& SVM & NUAA Imposter \\ &Gabor Wavelets + & & Yale Recaptured \\ & HOG & & Print-Attack \\ \hline

IDIAP~\cite{chakka:2011} & LBP & $\chi$ $^{2}$ & Print-Attack \\ \hline

UOULU~\cite{chakka:2011} & LBP & SVM & Print-Attack \\ \hline

CASIA~\cite{chakka:2011} & RASL + GMM +&&\\& Haar wavelets & LR & Print-Attack \\ \hline

Tirunagari {\it et al.}~\cite{tirunagari:2015} & DMD + LBP & SVM & Print-Attack \\ \hline

CASIA~\cite{chingovska:2013} & LBP + 1D-FFT + HMOF &&\\&  + Motion Correlation & SVM & Replay-Attack \\ \hline

LNMIIT ~\cite{chingovska:2013} & LBP + GMM + 2D-FFT & SVM & Replay-Attack \\ \hline

Feng \textit{et al.}~\cite{feng2016integration}	& HSC + Optical Flow & NN & Replay-Attack \\ \hline

Boulkenafet \textit {et al.}~\cite{boulkenafet2016face}	& LPQ + Color & SVM & Casia Face Anti-Spoofing \\ \hline

Kose and Dugelay~\cite{kosedugelay:2014} & LBP + Variational Retinex & SVM & Kose and Dugelay's \\ \hline

Menotti {\it et al.}~\cite{menotti:2015} & DNN & CNN & 3D Mask Attack \\ \hline

Feng \textit{et al.}~\cite{feng2016integration}	& HSC + Optical Flow & NN & 3D Mask Attack \\ \hline

Kim \textit{et al.}~\cite{kimahn:2016}	& MLBP + GLCM + IDA & SVM & MSU-MFSD \\ \hline

\end{tabular}
\end{table*}

In general, existing works seem to be going towards data set tuning ({\it i.e.} overfitting) instead of designing more effective and flexible solutions. This is corroborated by the works of Pereira {\it et al.}~\cite{de2013can} and Pinto {\it et al.}~\cite{pinto2015face}, which show initial cross-data set performance analyses using Casia Face Anti-Spoofing and Replay-Attack data sets. Different methods were evaluated by Pereira {\it et al.}~\cite{de2013can}, and Tables~\ref{tab24} and~\ref{tab25} show the results for the two most interesting ones, respectively, (1) Motion~Correlation+MLP; and (2) LBP-TOP+SVM. While LBP-TOP+SVM presents the best performance in experiments within a data set, Motion~Correlation+MLP performs better in experiments across different data sets, which seems to indicate that not necessarily the best performing works for a specific data set, like the ones shown in Table~\ref{tab11}, are actually the best countermeasures. On the other hand, they probably miss in terms of generalization power. Similar results can be also found in Pinto {\it et al.} work~\cite{pinto2015face}. Therefore, countermeasures with good performance in cross-data set experiments -- in the absence of a truly challenging data set -- are expected to be more effective in real world scenarios. Current countermeasures, however, hardly beat a random classifier ({\it i.e.} 50\% HTER).

\begin{table*}[!ht]
\caption{Cross-data set results (HTER) for Motion Correlation+MLP, as presented by Pereira \textit{et al.}[67].}\nocite{de2013can}
\label{tab24} 
\centering
\begin{tabular}{|l|c|c|}
\hline 
\diagbox{\textbf{Train}}{\textbf{Test}} & \textbf{Casia Face Anti-Spoofing} & \textbf{Replay-Attack} \\ \hline
Casia Face Anti-Spoofing & 30.33\% & 50.25\% \\ \hline
Replay-Attack & 48.28\% & 11.79\% \\ \hline
\end{tabular}
\end{table*}

\begin{table}[!ht]
\caption{Cross-data set results for LBP-TOP+SVM in HTER, as presented by Pereira \textit{et al.} [67].}\nocite{de2013can}
\label{tab25} 
\centering
\begin{tabular}{|l|c|c|}
\hline 
\diagbox{\textbf{Train}}{\textbf{Test}} & \textbf{Casia Face Anti-Spoofing} & \textbf{Replay-Attack} \\ \hline
Casia Face Anti-Spoofing & 23.75\% & 50.64\% \\ \hline
Replay-Attack & 61.33\% & 8.51\% \\ \hline
\end{tabular}
\end{table}

\subsection{Future perspectives}

Given the actual state of researches in face spoofing detection and the observed trends, we would like to point some future directions that could help other authors to address challenges that still need to be solved.

First, although texture-based solutions imported from face recognition systems have the best results in experiments within a data set, their performance rapidly degrade in experiments across different data sets~\cite{de2013can}. Thus, designing solutions specifically for spoofing detection like the initial works based on motion and reflectance seems to be a more promising way of achieving reasonable generalization. This topic has being understudied in the last years, but can find new stimuli in unexplored variations of deep learning that may benefit from this kind of information, such as long short-term memory networks~\cite{hochreiter:1997} and Fourier CNNs~\cite{pratt:2017}.

Second, other learning frameworks could be explored to offer a different perspective on how to solve this problem. Principles of lifelong~\cite{fischer:2000} and transfer learning~\cite{yu:2014} have not been explored so far. Such techniques would allow incorporating new samples into an existing model at any time, making it more flexible to cover further attacks in the future without retraining the entire classifier. In addition, clustering approaches~\cite{CORNUEJOLS201881} may be an option to analyze massive amounts of data in unsupervised or semi-supervised ways and could help to eventually discover unknown attacks without exhaustive manual annotation.

Third, a large web collected corpus for spoofing detection in uncontrolled scenarios would give an immediate boost to this field and would reduce overfitting problems related to data sets and/or attack types. More than that, this corpus could be created as extension of existing wild face recognition databases, such as Labeled Faces in the Wild~\cite{learnedmiller:2016}, to allow evaluating both recognition and spoofing detection simultaneously. To this end, one may search the web looking for images of individuals from of a chosen face recognition data set containing printed faces or even elaborate attacks like silicone masks and makeup disguises.

Finally, multimodal biometric systems are less likely to be spoofed as impostors, since one has to forge multiple biometric features at the same time. For this reason, different works addressed the impostor problem by combining two or more human characteristics~\cite{akhtar:2012,biggio:2012,johnson:2010,rodrigues:2009,rodrigues:2010,farmanbar:2017,singh:2017}. With this in mind, facial biometrics can be seen as a special case, since multimodality can take advantage of multiple facial properties ({\it e.g.,} texture, shape and temperature) to avoid spoof attacks. Nowadays, different commercially available devices are able to capture color, depth and infrared images simultaneously at a reasonable price. These devices could be used to enhance current countermeasures, and possibly make them practicable in industrial applications~\cite{litomisky2012consumer}.

\section{Conclusion}\label{sec:conclusion}

In this survey, we presented a compilation of face spoofing detection works over the past decade, as well as, a thoroughly numerical and qualitative analysis. Spoofing attacks persist to be a security challenge for face biometric systems, and there were much effort in the field to find robust methods. However, all these efforts have been following the same recipe, not favoring breakthroughs in the field. Many works of face spoofing detection give emphasis on 2D attacks by presenting printed photos or replaying recorded videos, and 3D attacks have been recently studied due to the technological advancements in 3D printer and reconstruction. Although perfect results on public data sets have been achieved by many works, there is a considerable gap to move from academic researching to real-world applications in a effective way. As such, it is expected that researchers concentrate efforts to create more difficulty data sets and more unbiased evaluation methods, henceforth.

\section*{Acknowledgments}
J. P. Papa is grateful to FAPESP grants \#2013/07375-0, \#2014/12236-1, and \#2016/19403-6, as well as CNPq grant \#306166/2014-3.

\section*{References}\label{sec:reference}

\bibliographystyle{ieeetr}

  \end{document}